\documentclass{arxiv_article}

\usepackage{microtype}
\usepackage{graphicx}
\usepackage{subcaption}
\usepackage{booktabs}
\usepackage{hyperref}

\usepackage{amsmath}
\usepackage{amssymb}
\usepackage{mathtools}
\usepackage{amsthm}
\usepackage{multirow}
\usepackage{xspace}
\usepackage{fancyvrb}
\usepackage{array}
\usepackage{tabularx}
\usepackage{ragged2e}

\usepackage{booktabs}
\usepackage{colortbl}
\usepackage{mathtools}
\usepackage{xcolor}

\usepackage{adjustbox}
\usepackage{nicematrix}
\usepackage{makecell}
\usepackage{graphicx}
\usepackage{enumitem}
\usepackage{wrapfig}
\usepackage{float}
\usepackage{fvextra}
\usepackage{siunitx}

\usepackage[most]{tcolorbox}
\usepackage{listings}
\definecolor{keywordcolor}{rgb}{0.0, 0.1, 0.6}   
\definecolor{tacticcolor}{rgb}{0.0, 0.1, 0.6}    
\definecolor{commentcolor}{rgb}{0.1, 0.5, 0.1}   
\definecolor{symbolcolor}{rgb}{0.0, 0.1, 0.6}    
\definecolor{sortcolor}{rgb}{0.0, 0.1, 0.6}      
\definecolor{attributecolor}{rgb}{0.7, 0.1, 0.1} 

\lstnewenvironment{leancode}{
\lstset{
    language=lean,
    breaklines,
}
}{}

\tcbset{
    commonstyle/.style={
        enhanced,
        breakable,
        boxrule=1pt,
        arc=2mm,
        fonttitle=\bfseries\sffamily,
        before skip=1.5em,
        after skip=1.5em,
    }
}
\newtcolorbox{databox}[2][]{
    commonstyle={#2},
    colback=blue!2!white,
    colframe=blue!50!black,
    title=#2,
    #1 
}

\newtcolorbox{codebox}[2][]{
    commonstyle={#2},
    colback=gray!5!white,
    colframe=gray!60!black,
    fontupper=\small\ttfamily, 
    title=#2,
    #1
}

\newtcolorbox{resultbox}[2][]{
    commonstyle={#2},
    colback=teal!2!white,
    colframe=teal!50!black,
    coltitle=white,
    title=#2,
    #1
}
\theoremstyle{plain}

\theoremstyle{definition}

\theoremstyle{remark}

\usepackage{wrapfig}
\usepackage{float}

\definecolor{lightpink}{rgb}{1,0.9,0.9}
\definecolor{lightblue}{rgb}{0.8,0.9,1.0}

\newcolumntype{C}[1]{>{\centering\arraybackslash}p{#1}}
\definecolor{oatpink}{HTML}{FDE2E4}
\definecolor{mypurple}{HTML}{4A148C}

\renewcommand\affiliationformat[2][]{ {\footnotesize $^{#1}$#2} }
\title{\LARGE TheoremForge: Scaling up Formal Data Synthesis with Low-Budget Agentic Workflow}
\author[1]{Yicheng Tao}
\author[1, 2, 3]{Hongteng Xu\thanks{Corresponding author. Email: hongtengxu@ruc.edu.cn}}
\affiliation[1]{Gaoling School of Artificial Intelligence, Renmin University of China}
\affiliation[2]{Beijing Key Laboratory of Research on Large Models and Intelligent Governance}
\affiliation[3]{Engineering Research Center of Next-Generation Intelligent Search and Recommendation, MOE}

\begin{document}
\abstract{The high cost of agentic workflows in formal mathematics hinders large-scale data synthesis, exacerbating the scarcity of open-source corpora. To address this, we introduce \textbf{TheoremForge}, a cost-effective formal data synthesis pipeline that decomposes the formalization process into five sub-tasks, which are \textit{statement formalization}, \textit{proof generation}, \textit{premise selection}, \textit{proof correction} and \textit{proof sketching}. By implementing a \textit{Decoupled Extraction Strategy}, the workflow recovers valid training signals from globally failed trajectories, effectively utilizing wasted computation. Experiments on a 2,000-problem benchmark demonstrate that TheoremForge achieves a Verified Rate of 12.6\%, surpassing the 8.6\% baseline, at an average cost of only \textbf{\$0.481} per successful trajectory using Gemini-3-Flash. Crucially, our strategy increases data yield by \textbf{1.6$\times$} for proof generation compared to standard filtering. These results establish TheoremForge as a scalable framework for constructing a data flywheel to train future expert models. Our code is available \href{https://github.com/timechess/TheoremForge}{here}.}

\maketitle

\section{Introduction}

Agentic workflows based on large language models have recently achieved significant advancements in the domain of formal mathematics. Compared to traditional single-turn interaction paradigms, agentic workflows demonstrate superior performance and robustness in autoformalization and automated theorem proving tasks~\citep{varambally2025hilbertrecursivelybuildingformal,chen2025seedprover15masteringundergraduatelevel,chen2025seedproverdeepbroadreasoning,xu-etal-2025-aristotle,wang2025ariaagentretrievaliterative}. This success is largely attributed to targeted strategy designs and test-time scaling techniques. However, the scarcity of training data hinders further development in this field. The diverse task requirements within workflows and the inherent data scarcity of the formalization domain exacerbate this challenge. Many current studies employ proprietary data for training while open-source formalization data remain insufficient in terms of quantity and diversity. Synthesizing training data suitable for formal reasoning models at a low cost and large scale has therefore become a critical research question.

\begin{wrapfigure}{r}{6cm}
    \centering
    \vspace{-26pt}
    \includegraphics[width=\linewidth]{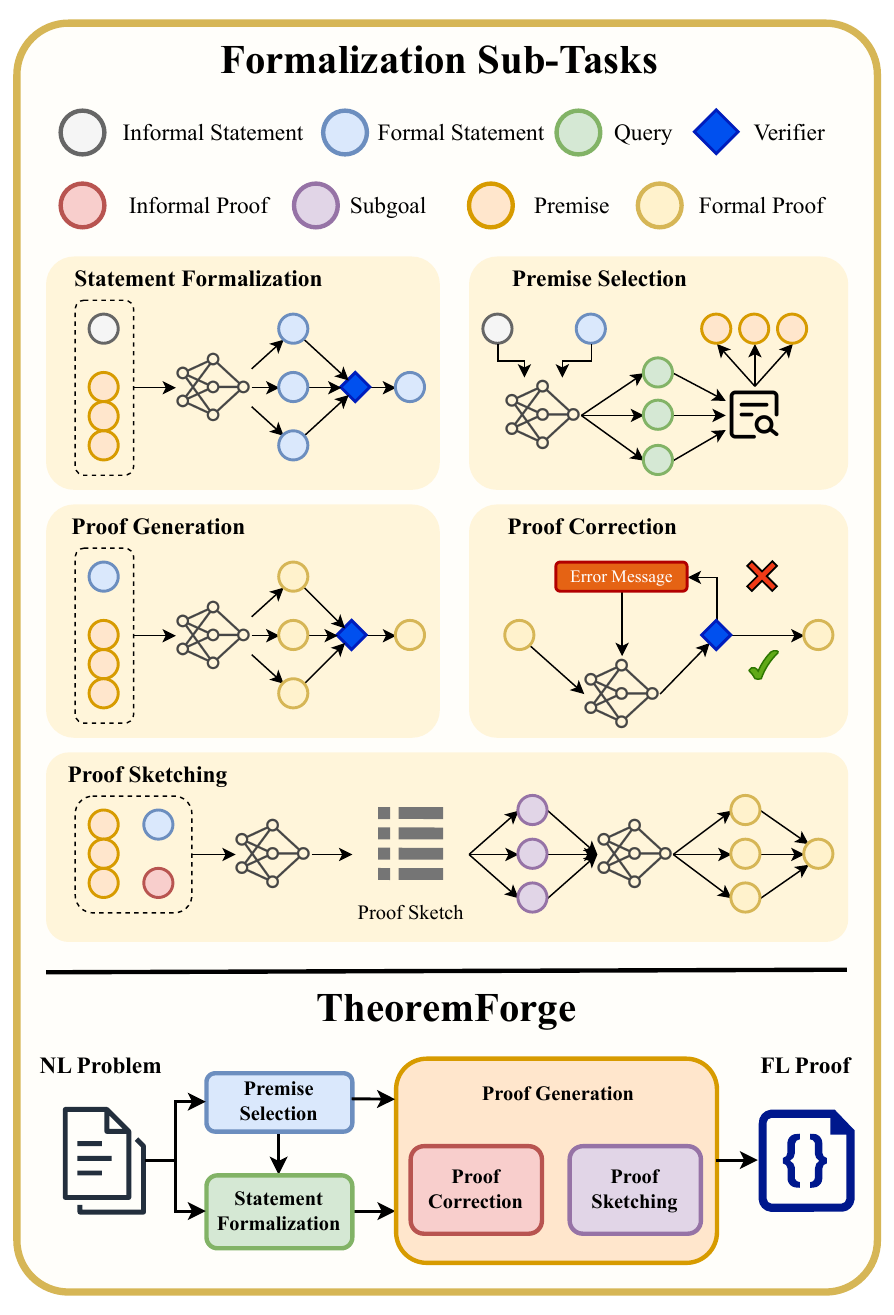}
    \caption{Schematic overview of the formalization workflow and its five sub-tasks.}
    \label{fig:workflow}
\end{wrapfigure}

Current formalization tasks are primarily categorized into Autoformalization and Automated Theorem Proving. A complete formalization workflow should be capable of translating natural language mathematical propositions into semantically consistent formal propositions and generating compilable proofs. We analyze common and effective strategy designs in existing methods to extract five sub-tasks (see Figure~\ref{fig:workflow}). These include statement formalization~\citep{gao2025herald,chen2025reform,wang2025ariaagentretrievaliterative}, proof generation~\citep{kimina_prover_2025,lin2025goedelproverv2,xin2025scaling,xin2025deepseekproverv,ren2025deepseekproverv2advancingformalmathematical}, premise selection~\citep{gao-etal-2024-semantic-search,asher2025leanexploresearchenginelean,liu2025rethinking,wang2025ariaagentretrievaliterative}, proof correction~\citep{lin2025goedelproverv2,chen2025seedproverdeepbroadreasoning}, and proof sketching~\citep{jiang2023draft, varambally2025hilbertrecursivelybuildingformal,chen2025seedprover15masteringundergraduatelevel,chen2025seedproverdeepbroadreasoning}. Optimizing these sub-tasks individually provides a modular path toward enhancing the global performance of the formalization pipeline. The open-source community currently lacks training data for tasks other than statement formalization and proof generation. Furthermore, the absence of expert models for the remaining sub-tasks necessitates the use of general-purpose large models as substitutes.

This paper focuses on data synthesis for these five sub-tasks and presents TheoremForge. This complete formalization workflow covers all the aforementioned tasks and serves as a data synthesis pipeline that achieves a favorable cost-performance trade-off. We utilize Gemini-3-Flash as the reasoning model to attain an average cost for a successful trajectory of \$0.481. Existing formalization workflows aim primarily to improve success rates and incur extremely high test-time scaling costs~\citep{varambally2025hilbertrecursivelybuildingformal,chen2025seedprover15masteringundergraduatelevel,chen2025seedproverdeepbroadreasoning}. These high costs manifest as large-scale parallel sampling and iterative subgoal decomposition which makes them unsuitable for large-scale data synthesis.

The contributions of this paper are summarized below.

\begin{itemize}[leftmargin=*]
    \item We design and open-source a formalization workflow capable of generating the full process from natural language mathematical propositions to formal proofs.
    \item The workflow serves as a low-cost data synthesis pipeline for formal training data which covers multiple sub-tasks and facilitates the training of agentic models.
    \item We evaluate the performance and cost of common formalization strategies under low inference budgets to provide a benchmark for future research.
\end{itemize}

\section{Preliminaries and Notations}

We formally define the five sub-tasks and their data formats. For concrete data instances, please refer to Appendix~\ref{appendix:data_example}.

\textbf{Statement Formalization} translates an informal statement $S_N$ into a formal specification $S_F$ conditioned on retrieved premises $P$. While syntax is compiler-verified, semantic equivalence is often evaluated via LLM-as-Judge~\cite{chen2025reform,gao2025herald} to ensure the mathematical meaning is preserved. Training samples are triplets $(S_N, P, S_F)$, where $S_F$ is both syntactically and semantically valid.

\textbf{Proof Generation} aims to generate a complete formal proof $R_F$ from $S_F$ and $P$ that passes the kernel check. Unlike translation, the correctness of $R_F$ is deterministically verifiable, providing an objective reasoning signal. Training samples are triplets $(S_F, P, R_F)$.

\textbf{Premise Selection} identifies pertinent definitions and theorems $P$ from a library~\cite{yang2023leandojo} using natural language queries $Q$ based on $S_N/S_F$. Effective selection reduces context noise and prevents model overwhelm. Training samples are triplets $(S_N/S_F, Q, P)$ where premises directly contribute to a successful downstream formalization or proof.

\textbf{Proof Correction} leverages compiler error messages $E$ to repair a failed proof $R_F$ into a successful version $R_F'$~\cite{lin2025goedelproverv2}. This task emphasizes iterative refinement and precise fault localization, significantly enhancing efficiency compared to re-sampling~\cite{chen2025seedproverdeepbroadreasoning}. Training samples are triplets $(R_F, E, R_F')$.

\textbf{Proof Sketching} decomposes complex theorems into manageable intermediate subgoals. Given $S_F$, $P$, and a natural language proof $R_N$, the model generates a compilable sketch $R_S$ as a high-level reasoning roadmap~\cite{varambally2025hilbertrecursivelybuildingformal}. Training samples are quadruplets $(S_F, P, R_N, R_S)$.

\begin{figure*}[t]
    \centering
    \includegraphics[width=\linewidth]{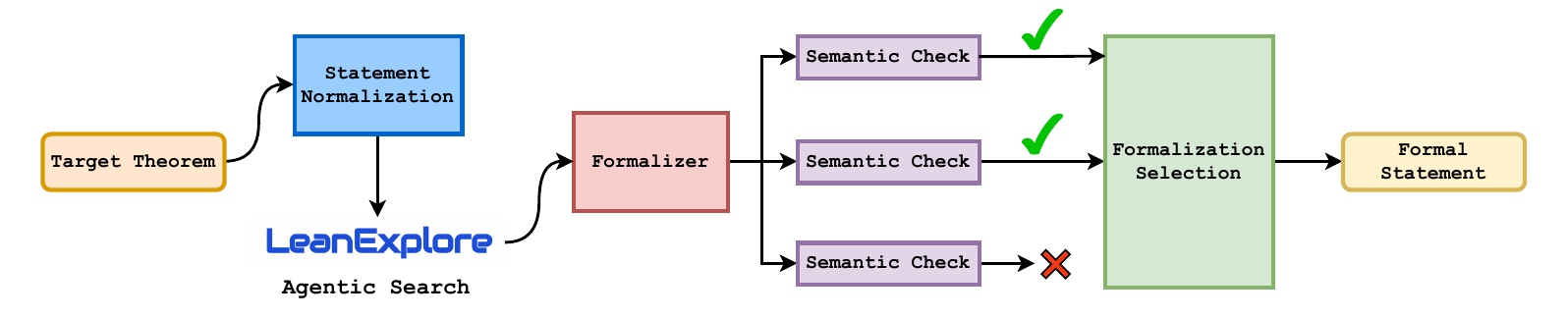}
    \caption{\textbf{Statement Formalization Workflow.} The pipeline consists of three phases: 1) \textbf{Information Preprocessing} for normalization and definition retrieval; 2) \textbf{Expert Model Sampling} for candidate generation and compilation checks; and 3) \textbf{Statement Filtering} for semantic verification via LLM-as-Judge and final selection.}
    \label{fig:statement_formalization}
\end{figure*}

\section{System Design}

In this section, we present the architecture of TheoremForge. While the system functions as a robust solver through two primary stages defined as \textbf{Statement Formalization} and \textbf{Proof Generation}, its overarching design objective is to serve as a high-efficiency data synthesis pipeline. We explicitly architect the workflow to support a \textbf{Decoupled Extraction Strategy} which maximizes data yield by independently validating and harvesting intermediate outputs from every inference trajectory. We detail the execution logic of the formalization and proving stages followed by the specific mechanisms for data extraction. See Appendix~\ref{appendix:config} for details.

\subsection{Statement Formalization}

The system must derive a syntactically correct and semantically aligned formal statement from the input prior to generating a proof. The workflow for this process is illustrated in Figure~\ref{fig:statement_formalization}.

\textbf{Information Preprocessing.} Informal input statements exhibit diverse characteristics in both type and form. This heterogeneity often compels the formalization process to encompass ancillary tasks such as supplying implicit premises, solving for numerical answers, and interpreting ambiguous semantics. These factors can adversely affect model performance. We first employ a general-purpose LLM to denoise the input statement before engaging the expert formalizer to address these challenges. We subsequently implement an agentic premise selection strategy leveraging the open-source search engine LeanExplore~\cite{asher2025leanexploresearchenginelean}. The model generates search queries $Q$ based on the normalized statement $S_N$ and selectively incorporates valid premises $P$ from the retrieval results into the context.

\textbf{Statement Filtering.} The workflow imposes a series of filtering procedures on the statements $S_F$ generated by the expert model to enhance the quality of formalization. These procedures encompass both syntactic and semantic verifications. We retain only those candidates that successfully pass both compilation checks and semantic evaluations based on the LLM-as-Judge paradigm. The optimal statement is then selected as the final output of the formalization phase. This process effectively eliminates distractors that exhibit obvious semantic inconsistencies although it does not guarantee the absolute correctness of the formalization.

\subsection{Proof Generation}

\begin{figure*}[t]
    \centering
    \includegraphics[width=\linewidth]{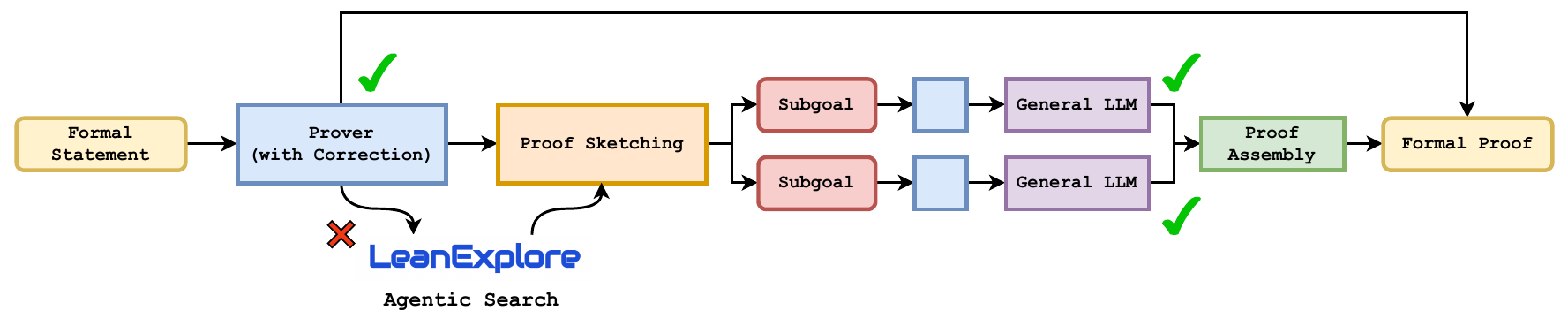}
    \caption{\textbf{Proof Generation Workflow.} The process comprises two phases. 1) \textbf{Expert Model Sampling} is where an expert model is invoked to generate a proof. A general-purpose LLM attempts to rectify the errors if all candidates fail. 2) \textbf{Subgoal Decomposition} is where a proof sketch is generated and the problem is decomposed into subgoals.}
    \label{fig:proof_generation}
\end{figure*}

We employ the workflow illustrated in Figure~\ref{fig:proof_generation} to search for a valid proof if the system successfully formalizes the statement. The design of this workflow draws inspiration from Hilbert~\cite{varambally2025hilbertrecursivelybuildingformal} and principally adopts strategies involving iterative proof refinement driven by compiler feedback together with subgoal decomposition based on proof sketches.

\textbf{Iterative Proof Refinement.} The workflow captures compiler feedback from failed attempts by either the expert or general-purpose LLM to guide repairs. To prevent context pollution and ensure synthesized data remains consistent with our formal task definition, we restrict the model's visibility strictly to the incorrect code $R_F$ and current error message $E$, excluding interaction history. This stateless design ensures that each refinement step can be extracted as an independent, high-quality $(R_F, E, R_F')$ correction pair, effectively maximizing data yield from globally failed trajectories. We maintain scalability by capping iteration rounds to balance reasoning depth with computational expenditure.

\textbf{Subgoal Decomposition.} The workflow proceeds to the subgoal decomposition phase should the proof remain incomplete relying solely on the expert model and proof correction. This stage initiates with an agentic premise selection process analogous to that used in statement formalization yet is distinguished by the fact that the model generates retrieval queries based specifically on the formal statement $S_F$. The premises $P$ obtained thereby serve as the shared context for all subsequent subgoal proof attempts. The workflow builds upon this foundation by prompting the general-purpose LLM to generate a comprehensive step-by-step natural language proof $R_N$ before attempting to construct a formal proof sketch $R_S$. The embedded subgoals are extracted for parallel resolution via a generation sequence that utilizes the expert model followed by the general-purpose LLM provided the sketch successfully passes compilation. These subgoals are synthesized with the original sketch to constitute the complete formal proof $R_F$ upon the successful verification of all subgoals.

\subsection{Decoupled Data Extraction Strategy}

The efficiency of TheoremForge as a data synthesis pipeline stems from its ability to isolate valid training signals even when the global reasoning trajectory fails. Unlike traditional approaches that filter solely for complete proofs, we implement a decoupled extraction strategy that harvests high-quality training samples for individual sub-tasks based on local verification signals. This approach ensures that intermediate successes such as a correct formal statement or a verified subgoal are preserved regardless of the final outcome. We delineate the sourcing logic for each specific data type below.

\textbf{Statement Formalization Data.} We extract valid triplets $(S_N, P, S_F)$ from all trajectories where the formal statement $S_F$ passes the compilation check and the semantic check. This process is independent of the subsequent proof generation success and allows us to harvest formalization data at a high rate.
    
\textbf{Proof Generation Data.} We extract verified proofs from two distinct sources. The first source is the successful trajectories of the main theorems. The second source comprises the individual subgoals that are successfully proved during the decomposition phase even if the parent theorem remains unproved. This strategy significantly expands the dataset volume as a single failed trajectory often yields multiple valid lemma-proof pairs from its subgoals.

\textbf{Premise Selection Data.} We construct positive samples by strictly filtering for premises that explicitly appear in the successfully generated code, including successful subgoals in a failed trajectory. This ensures that the selected premises are not merely retrieved but are essential for the derivation. Hard negatives can be easily generated using the invalid queries and unused retrieval results.

\textbf{Proof Correction Data.} The iterative refinement module provides a rich source of correction pairs $(R_F, E, R_F')$. We capture every instance within the refinement loop where the model successfully fixes a compilation error regardless of whether the final proof is completed. This mechanism effectively converts intermediate failures into valuable debugging data.
    
\textbf{Proof Sketching Data.} We capture proof sketches generated during the subgoal decomposition phase. While we collect all sketches that pass the compilation check to ensure structural diversity, we note that only those sketches that result in a fully closed proof are guaranteed to be logically sound. These verified sketches constitute the high-quality core of the dataset while the remaining compilable sketches serve as supplementary data or hard negative cases for downstream training.
\section{Experiments}

We conduct a comprehensive evaluation of TheoremForge to validate its efficacy as a scalable data synthesis pipeline for formal mathematics. Our experiments focus on assessing the capabilities of general-purpose LLMs within this workflow. We place particular emphasis on the trade-off between reasoning performance and computational cost to ensure the feasibility of large-scale data generation.

We organize our experimental analysis around four pivotal research questions.

\begin{itemize}[leftmargin=*]
    \item \textbf{RQ1:} How do different general-purpose foundation models compare in utilizing the workflow for formalization, and which model offers the optimal balance between capability and efficiency?
    \item \textbf{RQ2:} Is TheoremForge a cost-effective solution for large-scale data synthesis, and what are the specific resource expenditures associated with successful trajectories?
    \item \textbf{RQ3:} How robust is the workflow across diverse mathematical domains and difficulty levels, and what characterizes the distribution of the synthesized data?
    \item \textbf{RQ4:} How do individual agentic modules contribute to the overall success of the workflow, and what are the characteristic bottlenecks that lead to failed trajectories?
\end{itemize}

In the following subsections, we first detail the experimental setup including benchmarks and baselines before presenting empirical results that address these research questions sequentially.

\subsection{Experimental Setup}\label{sec:setup}

\textbf{Datasets.} We construct a small-scale (100 problems) and a large-scale (2,000 problems) evaluation set by sampling uniformly from \textbf{DeepMath}~\citep{deepmath}, which comprises 103K challenging problems, and \textbf{DeepTheorem}~\citep{zhang2025deeptheoremadvancingllmreasoning}, which contains 121K IMO-level proof-based theorems.

\textbf{Baseline.} Due to the lack of open-source end-to-end workflows, we establish a baseline by directly applying expert models without agentic orchestration: ReForm-32B~\citep{chen2025reform} for formalization and Goedel-Prover-32B~\citep{lin2025goedelproverv2} for proving. To ensure fair comparison, these models also serve as the underlying expert modules in TheoremForge, with a consistent sampling budget of $n=4$.\footnote{The workflow will make extra expert calls. However, the improvement in sampling budget can be marginal.}

\textbf{Metrics.} We evaluate the workflow across three metrics: (1) \textbf{Formalization Rate (FR)}: the percentage of successfully compiled statements; (2) \textbf{Proof Rate (PR)}: the percentage of valid formal proofs generated; and (3) \textbf{Verified Rate (VR)}: the percentage of problems judged semantically consistent by a majority vote of LLM verifiers. To ensure independence, for each problem, we exclude the original generator from the pool of 7 participating models and use the remaining 6 as judges\footnote{DeepSeek-V3.2 variants (with and without thinking) are treated as a single model entity. If either variant is the generator, both are excluded from the voting pool to ensure independence.}. See Appendix~\ref{appendix:semantic} for details.

\textbf{Environment.} We use Lean v4.19.0 to match the LeanExplore~\citep{asher2025leanexploresearchenginelean} database. 

For detailed configurations, see Appendix~\ref{appendix:config}.

\subsection{RQ1: Model Selection and Performance Analysis}

\begin{table*}[htbp]
\centering
\caption{Model performance comparison on the small-scale benchmark. Gemini-3-Pro and Gemini-3-Flash are tested with low thinking level. The tokens and cost are calculated excluding expert models. We define Expert Calls as the total number of sampling invocations by the expert model, where a pass@4 task is counted as four calls. \fbox{Best} and \textbf{Second Best} results are highlighted.}
\label{tab:model_comparison}
\resizebox{\textwidth}{!}{%
\begin{tabular}{l ccc ccc c c c c}
\toprule
\multirow{2}{*}{\textbf{Model}} & \multicolumn{3}{c}{\textbf{DeepMath}} & \multicolumn{3}{c}{\textbf{DeepTheorem}} & \textbf{Overall} & \textbf{Expert} & \textbf{Tokens} & \textbf{Cost} \\
\cmidrule(lr){2-4} \cmidrule(lr){5-7}
 & FR & PR & VR & FR & PR & VR & VR & Calls & (M) & (USD) \\
\midrule
GPT-5.2 & 74.0\% & 14.0\% & 14.0\% & 58.0\% & 16.0\% & 14.0\% & 14.0\% & 1372 & 3.41 & \$17.27 \\
Claude-Sonnet-4.5 & 88.0\% & \textbf{28.0\%} & \textbf{26.0\%} & \fbox{80.0\%} & \textbf{36.0\%} & \textbf{32.0\%} & \textbf{29.0\%} & 2072 & 5.12 & \$32.83 \\
Gemini-3-Pro (low) & 90.0\% & \fbox{32.0\%} & \fbox{30.0\%} & \textbf{76.0\%} & \fbox{42.0\%} & \fbox{36.0\%} & \fbox{33.0\%} & 1608 & 7.90 & \$65.21 \\
Gemini-3-Flash (low) & \fbox{94.0\%} & 26.0\% & 20.0\% & 72.0\% & 32.0\% & 26.0\% & 23.0\% & 1616 & 4.70 & \$6.94 \\
DeepSeek-V3.2-Thinking & 70.0\% & 10.0\% & 8.0\% & 54.0\% & 18.0\% & 18.0\% & 13.0\% & 900 & 3.09 & \$2.72 \\
DeepSeek-V3.2 & 68.0\% & 18.0\% & 14.0\% & 62.0\% & 22.0\% & 22.0\% & 18.0\% & 1100 & 3.14 & \$2.78 \\
Qwen-Max & 88.0\% & 10.0\% & 10.0\% & \textbf{76.0\%} & 22.0\% & 22.0\% & 16.0\% & 796 & 2.94 & \$2.45 \\
Baseline & \textbf{92.0\%} & 16.0\% & 14.0\% & \textbf{76.0\%} & 20.0\% & 16.0\% & 15.0\% & 736 & -- & -- \\
\bottomrule
\end{tabular}%
}
\end{table*}
We specifically investigate the performance of various general-purpose large language models acting as the reasoning core of TheoremForge to answer the first research question. The results presented in Table~\ref{tab:model_comparison} reveal significant disparities in both formalization capability and economic efficiency across different models.

\textbf{Effectiveness of the Agentic Workflow.} A comparison between the baseline and the agentic configurations highlights the substantial value added by the workflow. The baseline method employs the expert models directly and achieves a high Formalization Rate of 92.0\% on DeepMath. This indicates that the ReForm-32B model is proficient at generating syntactically correct Lean code. However, the verified rate of the baseline remains limited at 15.0\% overall. In contrast, workflows driven by capable reasoning models such as Gemini-3-Pro and Claude-Sonnet-4.5 significantly outperform the baseline with verified rates reaching 33.0\% and 29.0\% respectively. This empirical evidence suggests that the inclusion of premise selection, iterative proof refinement, and subgoal decomposition modules effectively converts static formalization capability into successful theorem proving outcomes.


\begin{figure*}[ht]
\centering
\begin{subfigure}[b]{.38\textwidth}
    \centering
    \includegraphics[height=5.4cm, keepaspectratio]{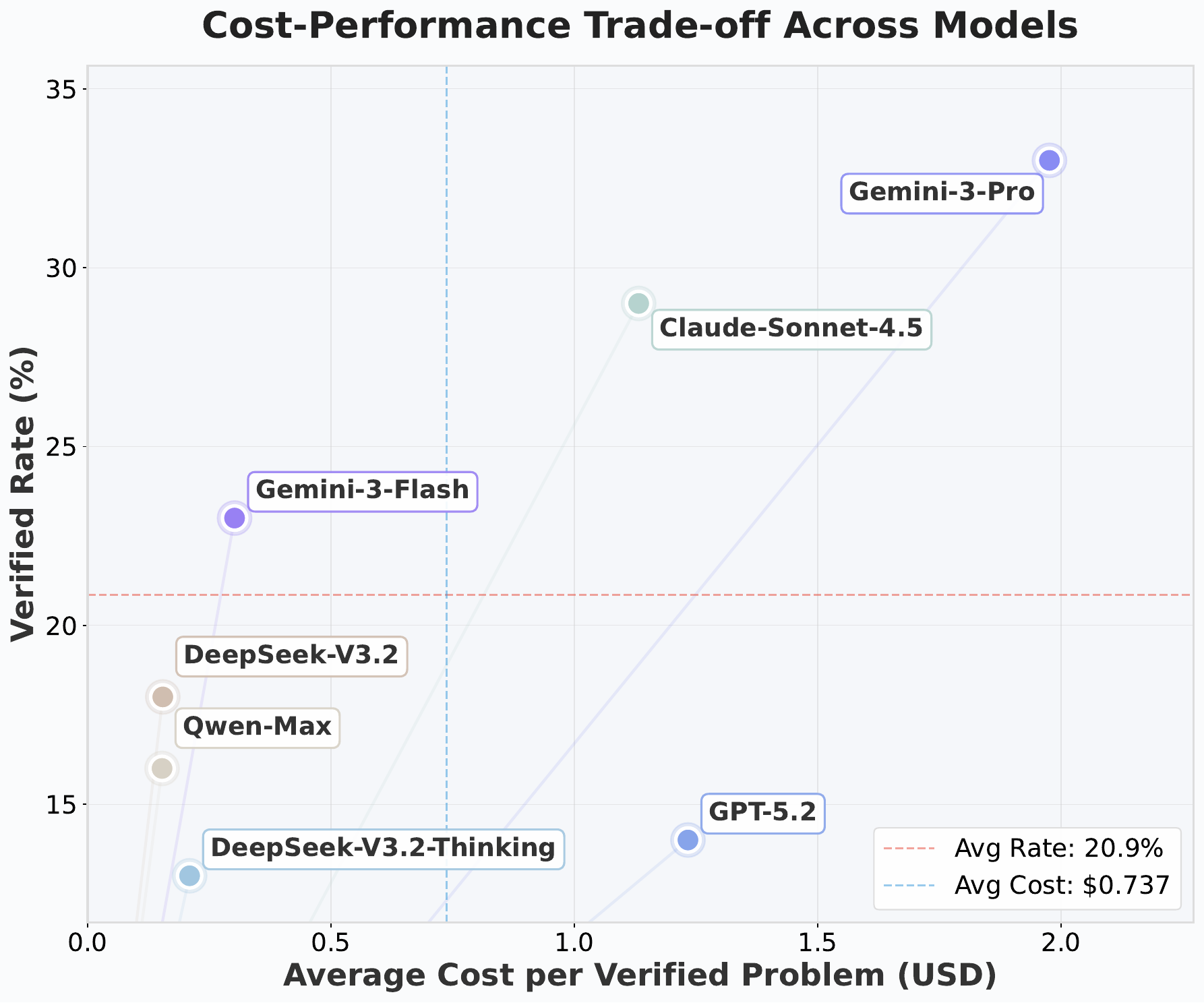}
    \caption{}
     \label{fig:cost_performance}
\end{subfigure}
\begin{subfigure}[b]{.58\textwidth}
    \centering
    \includegraphics[height=5.4cm, keepaspectratio]{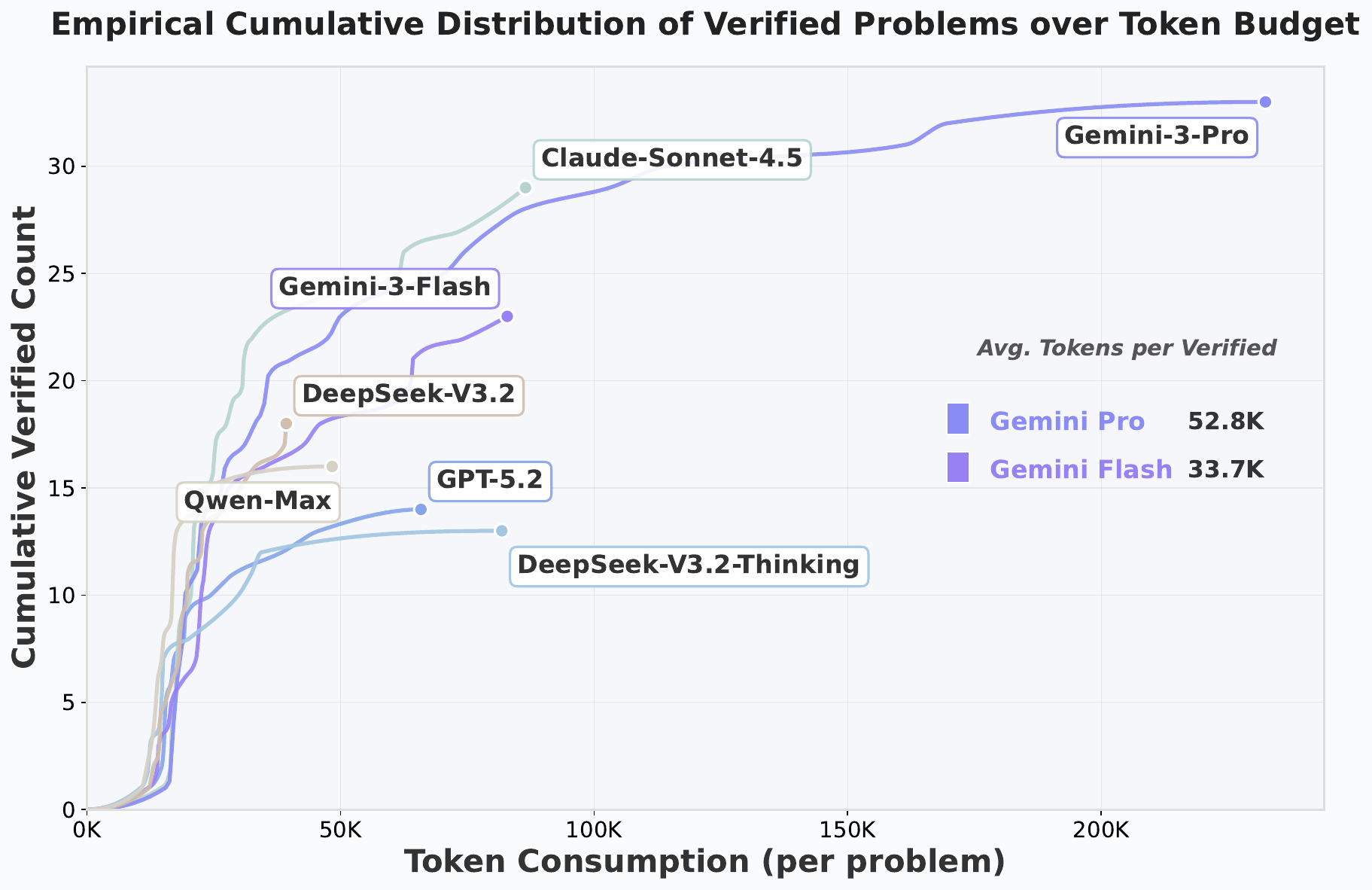}
    \caption{}
    \label{fig:token_performance}
\end{subfigure}
    \caption{(a) This figure illustrates the cost–performance comparison across different models, where models located in the upper-left region achieve a more favorable balance. The average cost is calculated by Total Cost / Number of Verified Problem. (b) This figure compares models by the cumulative number of verified problems as a function of per-problem token consumption, revealing their efficiency under limited computational budgets.}
\end{figure*}
\textbf{Cost-Performance Trade-off.} The feasibility of large-scale data synthesis depends critically on the balance between success rate and inference cost. We visualize this relationship in Figure~\ref{fig:cost_performance} which maps the verified rate against the average cost per verified problem. The plot reveals that Gemini-3-Flash occupies the optimal upper-left region of the Pareto frontier. This positioning indicates that it achieves high verification performance while maintaining minimal expenditure per successful instance. While Gemini-3-Pro delivers the highest raw performance, it is situated in the upper-right high-cost region and incurs a substantial total cost of \$65.21. Conversely, Gemini-3-Flash achieves 69.7\% of the performance of the Pro version but reduces the total cost by approximately 89.4\% to merely \$6.94. This also represents a significant efficiency advantage over Claude-Sonnet-4.5 which costs \$32.83. DeepSeek and Qwen appear in the bottom-left region as the most economical options but their low success rates render them unsuitable for generating high-quality training data.

\textbf{Mechanism of Collaboration.} We previously demonstrated the synergy of general-purpose LLM and expert model. Figure~\ref{fig:token_performance} further details this mechanism by plotting cumulative success against the token consumption of the general-purpose LLM. The extended tails observed for Gemini-3-Pro demonstrate that powerful reasoning models effectively leverage larger token budgets to raise the performance upper bound of the workflow. In contrast, less capable models tend to plateau early regardless of the available budget.


\subsection{RQ2: Scalability and Data Synthesis Efficiency}

We assess the scalability of TheoremForge by deploying the Gemini-3-Flash powered workflow on the large-scale benchmark comprising 2,000 problems. This experiment evaluates both the robustness of the system on unseen data and the economic feasibility of mass production.

\textbf{Large-scale Performance and Cost.} The results presented in Table~\ref{tab:gemini_flash_stats} demonstrate the superior performance of our workflow compared to the baseline expert models. Gemini-3-Flash achieves a verified rate of 12.60\% which surpasses the 8.60\% rate of the baseline. This improvement confirms that the agentic strategies remain effective when scaled to a larger and more diverse problem set. Most critically, the average cost for generating a successful verified trajectory is merely \$0.481. This is a manageable budget for academic and industrial research labs relative to the high value of formal mathematics data.

\begin{table}[htbp]
\centering
\caption{Comparison between TheoremForge (with Gemini-3-Flash) and Baseline on the large-scale benchmark.}
\label{tab:gemini_flash_stats}
\resizebox{.7\textwidth}{!}{%
\begin{tabular}{l c c c c c}
\toprule
\textbf{Method} & \textbf{FR} & \textbf{PR} & \textbf{VR} & \textbf{Expert Calls} & \textbf{Avg Cost} \\
\midrule
TheoremForge & 81.30\% & 14.45\% & 12.60\% &  24056 & \$0.4806 \\
Baseline & 72.25\% & 12.55\% &8.60\% & 13780 & --\\
\bottomrule
\end{tabular}%
}
\end{table}

\begin{figure}[htbp]
    \centering
    \includegraphics[width=.8\linewidth]{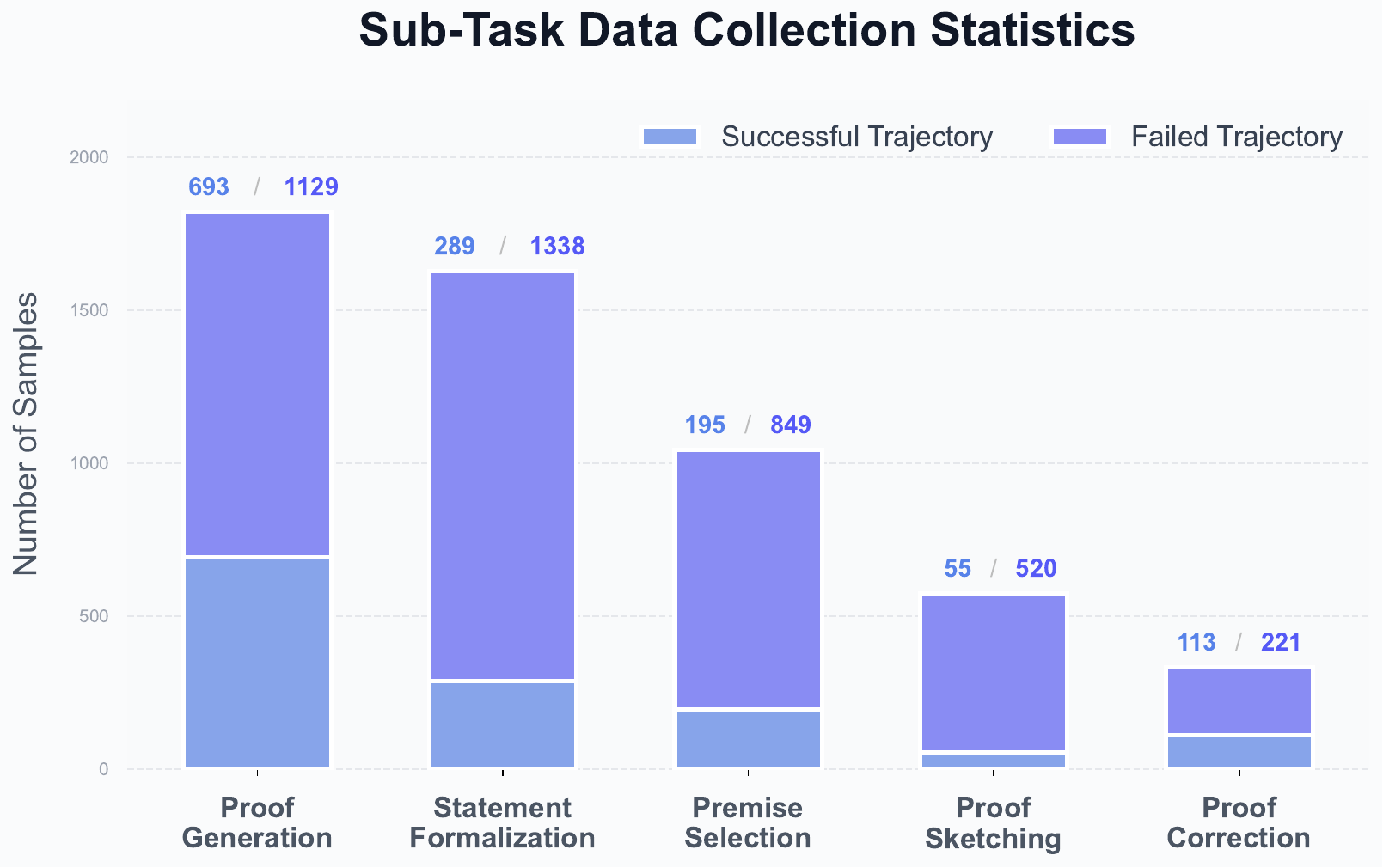} 
    \caption{Data collection statistics for the five sub-tasks. The stacked bars distinguish between samples derived from fully successful trajectories (bottom) and valid samples extracted from intermediate steps of failed trajectories (top). The latter reveals data yield improvements by decoupled extraction..}
    \label{fig:data_yield}
\end{figure}
\textbf{Efficiency of Decoupled Extraction.} The statistics presented in Figure~\ref{fig:data_yield} validate the success of our strategy in salvaging valuable data from globally failed inference trajectories. We observe that valid samples extracted from failed trajectories constitute the majority of the dataset for key tasks like statement formalization and proof generation. Specifically, the workflow harvests 1,338 valid formalization samples from failed runs which is more than four times the number obtained from successful trajectories. Similarly, the sub-task of proof generation benefits from subgoal decomposition as it yields 1,129 valid lemma proofs from failed trajectories compared to only 693 from successful ones. This result demonstrates that TheoremForge effectively maximizes the utility of computational resources by identifying and preserving local successes within broader failures.

\begin{figure*}[t]
    \centering
    \begin{subfigure}[b]{0.36\textwidth}
        \centering
        \includegraphics[height=5.5cm, keepaspectratio]{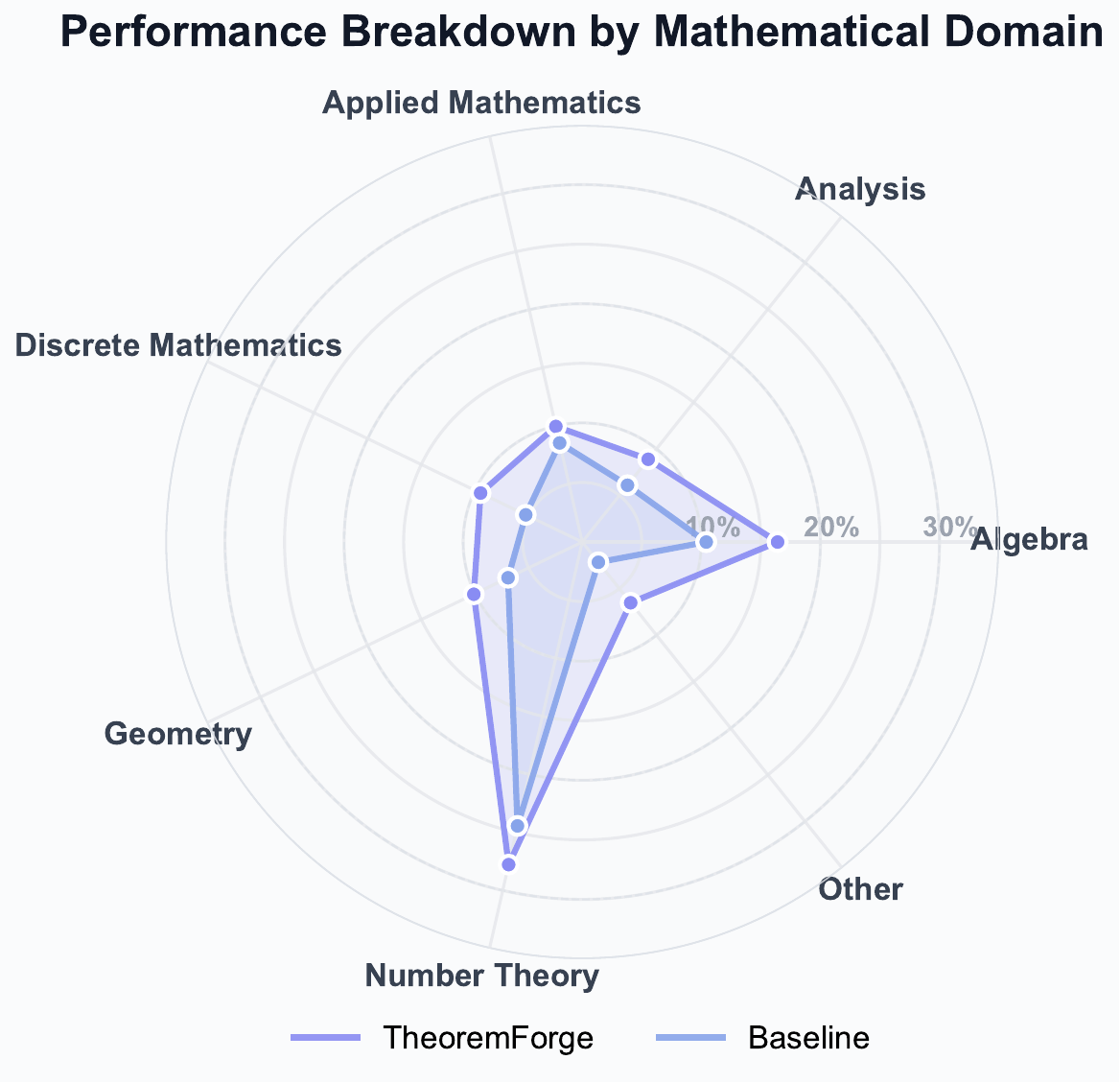} 
        \caption{}
        \label{fig:domain_radar}
    \end{subfigure}
    \hfill
    \begin{subfigure}[b]{0.62\textwidth}
        \centering
        \includegraphics[height=5.5cm, keepaspectratio]{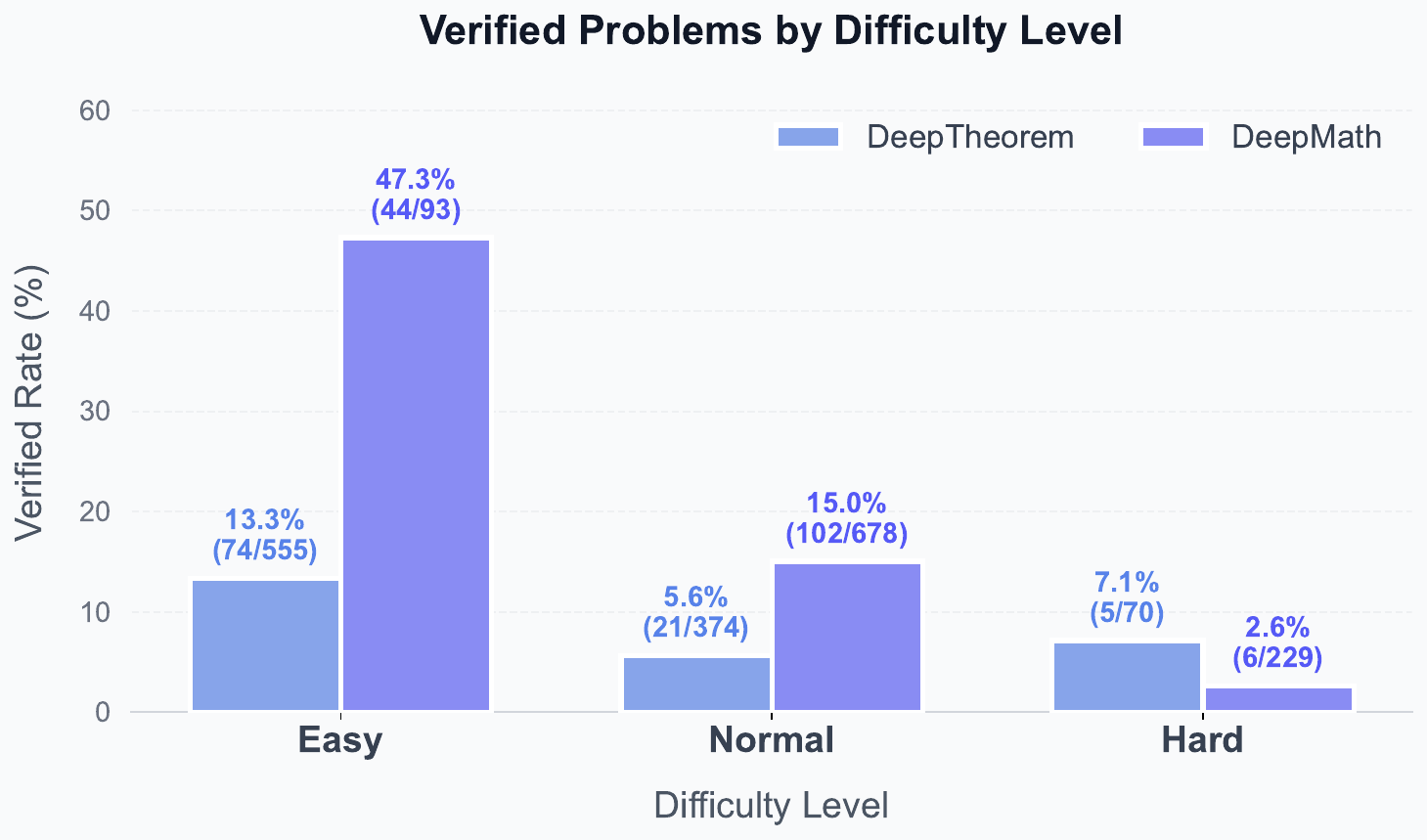} 
        \caption{}
        \label{fig:difficulty_bar}
    \end{subfigure}
    
    \caption{\textbf{Robustness Analysis across Domain and Difficulty.} (a) The radar chart illustrates the verified rate across five aggregated mathematical domains where TheoremForge consistently outperforms the baseline. (b) The bar chart details the verification statistics across three difficulty tiers. While success rates decline with complexity, the workflow successfully synthesizes a valuable set of high-difficulty training samples.}
    \label{fig:rq3_combined}
\end{figure*}
\textbf{Necessity of Specialized Training.} The relatively lower data yields observed in proof sketching and proof correction reveal the inherent limitations of current general-purpose LLMs in handling these specific tasks. The total volume of collected samples for proof correction is the lowest among all sub-tasks with only 334 instances combined. This scarcity indicates that general-purpose LLMs struggle to accurately diagnose and repair functional code based on compiler error messages. A similar trend is evident in proof sketching where the model often fails to generate logically sound high-level structures that lead to valid proofs. These findings underscore the critical need for training specialized expert models for these tasks. The data synthesized by our pipeline, though limited in quantity for these specific tasks, serves as a high-quality seed corpus to initiate the training of such experts.

\subsection{RQ3: Domain Robustness and Data Distribution}

We analyze the distribution of the synthesized data across diverse mathematical domains and difficulty levels to characterize the quality and training potential of the generated corpus. 

\textbf{Metadata Unification.} To facilitate a standardized analysis, we unify the heterogeneous metadata schemas from DeepTheorem and DeepMath. For domain classification, we aggregate specific tags into broad macro-categories including Algebra, Geometry, Analysis, Discrete Mathematics, Number Theory, Applied Mathematics and Other. We utilize the primary label for classification when a problem contains multiple domain tags. We further align the differing numerical difficulty scales into three tiers defined as Easy, Normal, and Hard. Specifically, we define Easy as difficulty values up to 6 for DeepTheorem and 4 for DeepMath. We classify values above 8 for DeepTheorem and 7 for DeepMath as Hard while the intermediate ranges are designated as Normal.

\textbf{Domain Robustness.} Figure~\ref{fig:domain_radar} illustrates the Verified rate across different mathematical fields. TheoremForge consistently surpasses the baseline in all categories and demonstrates a comprehensive coverage of formal mathematics. The workflow exhibits particular strength in Algebra and Number Theory where the verified rates are notably high. This performance suggests that the agentic strategies effectively handle symbolic manipulation and arithmetic properties which are foundational to formal reasoning. This broad domain coverage ensures that the synthesized dataset promotes semantic diversity.

\textbf{Adaptability to Difficulty.} Figure~\ref{fig:difficulty_bar} presents the distribution of verified problems across difficulty levels. We observe a natural inverse correlation between problem complexity and verified rate. The DeepMath dataset exhibits a high success rate of 47.3\% in the Easy category which provides a solid foundation of fundamental training data. DeepTheorem generally presents a greater challenge although the workflow successfully verifies a combined total of 11 Hard problems across both datasets. These successful high-difficulty samples are particularly valuable as they represent complex reasoning paths that contain rich logical structures. The resulting data distribution forms a natural curriculum that spans from abundant simple exercises to sparse but high-quality complex theorems.

\subsection{RQ4: Component Analysis and Failure Modes}\label{sec:rq4}
We analyze the results on large-scale benchmark to identify the performance gains from each module and categorize failure trajectories.

\begin{figure}[ht]
    \centering
    \includegraphics[width=\linewidth]{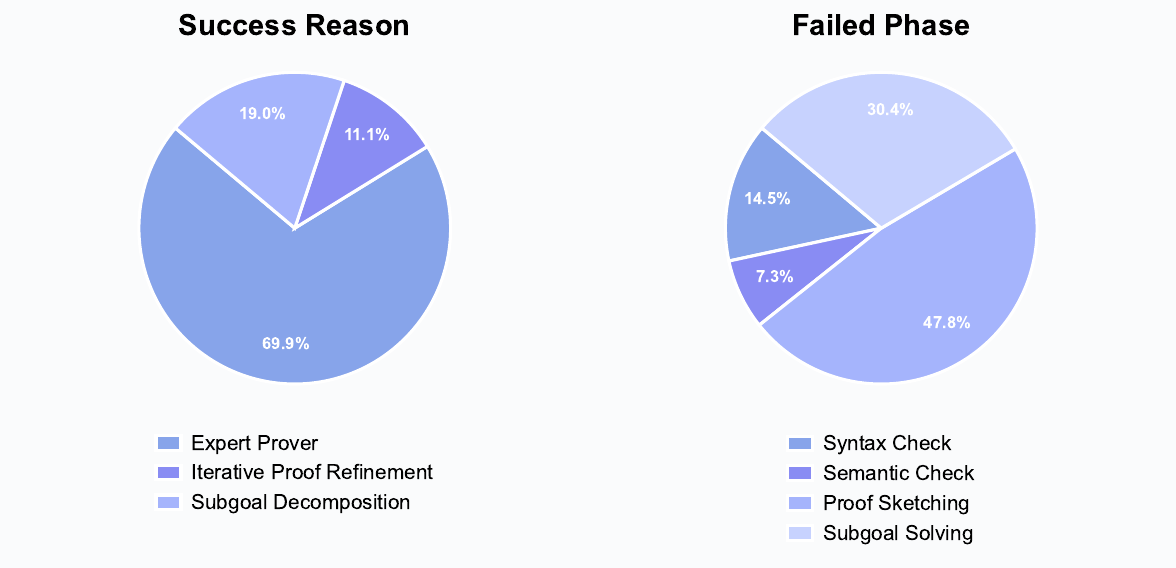}
    \caption{\textbf{Analysis of the Formalization Workflow.} (Left) Success reasons for verified proofs. (Right) Distribution of failed phases.}
    \label{fig:success_failure_analysis}
\end{figure}

\textbf{Module Contributions.} As shown in Figure~\ref{fig:success_failure_analysis} (Left), while the \textbf{Expert Prover} completes 69.9\% of tasks, agentic modules provide a 30.1\% boost: \textbf{Subgoal Decomposition} accounts for 19.0\%, tackling complex theorems beyond monolithic horizons, and \textbf{Iterative Refinement} contributes 11.1\% by repairing local tactical errors via compiler feedback.

\textbf{Failure and Bottleneck Analysis.} Figure~\ref{fig:success_failure_analysis} (Right) identifies \textbf{Proof Sketching} as the primary bottleneck (47.8\% of failures). Although the system generated 575 compilable sketches, many were logically unsound or led to unjustifiable subgoals, highlighting the challenge of ensuring mathematical viability over mere syntactic correctness. 

\textbf{Subgoal Dynamics and Efficiency.} \textbf{Subgoal Solving} is the second-largest failure category (30.3\%). Across 5,763 extracted subgoals ($\sim$10 per sketch), the expert model proved 1,372, with 161 recovered through refinement and 196 definitive failures. Crucially, 4,034 subgoals were bypassed by our \textit{early stop policy}, which terminates a sketch's remaining subgoals as soon as one fails. This mechanism significantly optimizes resources by pruning unviable trajectories. 

Detailed error taxonomies are in Appendix~\ref{appendix:error_analysis}.

\section{Related Works}

\subsection{Agentic Architectures}

Agentic workflows extend single-pass inference through iterative planning and tool integration which enables structured reasoning for complex tasks. Recent research emphasizes automating architecture design via agentic supernets to search for optimal multi-agent compositions~\citep{zhang2025aflow,wang2025evoagentx,icml2025_supernet,hu2025qualityflow}. Self-improvement mechanisms further allow agents to adaptively develop and maintain complex software libraries through continuous iteration~\citep{iclr2025_adaptive,wang2025dyflow}. Execution-aware approaches integrate runtime feedback or external solvers to refine solutions~\citep{ni-etal-2025-tree,yuksekgonul2025optimizing,singh2505agentic,nezhad2025symcode}. Theoretical frameworks for heterogeneous agents also investigate strategic collaboration to ensure rationality and truthful contribution among diverse components~\citep{icml2025_collaborative}. These advancements collectively support the shift towards modular systems where specialized agents collaborate to solve long-horizon problems.

\subsection{AI for Formal Mathematics}

The integration of large language models with formal methods is increasingly recognized as a prerequisite for building trustworthy AI agents~\citep{icml2025_trustworthy,wang2024theoremllama,yao-etal-2025-fans}. Contemporary systems in formal mathematics predominantly employ modular workflows that decompose theorem proving into coordinated sub-tasks. These systems leverage the Lean kernel for rigorous feedback and enable scalable exploration through specialized components~\citep{lin2025goedelproverv2,xin2025scaling,varambally2025hilbertrecursivelybuildingformal,chen2025seedprover15masteringundergraduatelevel,chen2025seedproverdeepbroadreasoning,hubert2025olympiad,xu-etal-2025-aristotle,hu2025hybridproveraugmentingtheoremproving,baba2025prover,ospanov2025apollo,kozyrev2025rocqstar,wang2025qdtsynth}. For autoformalization tasks, recent methods shift from purely semantic translation to process-driven approaches that exploit tactic structures and type checkers to guide stepwise formalization~\citep{jana2025proofbridgeautoformalizationnaturallanguage,lu2025processdriven,wang2025ariaagentretrievaliterative}.
\section{Conclusion and Discussion}

In this paper we present TheoremForge, a formal data synthesis pipeline to accumulate high-quality training data for downstream formalization tasks. It shows potential in establishing a data flywheel by iteratively training sub-task experts using the synthetic data. Here are some discussions about the insights and limitations of our work.

\textbf{Decoupling Reasoning from Coding.} Our workflow underscores the necessity of specialized agents over monolithic solvers. Since mathematical reasoning and formal coding require divergent cognitive capabilities, and specialized models often lag behind general-purpose ones in reasoning~\cite{jiang2025fate}, a collaborative paradigm that divides labor based on model strengths is more effective for complex formalization.

\textbf{Efficiency and Scaling.} To bridge the gap between vast natural language mathematics and scarce formal corpora, synthesis must optimize both quality control and cost-performance. We posit that the future of scalable formalization lies in deploying lightweight, specialized models on distinct sub-tasks. Running such models locally can maximize throughput and minimize computational overhead without relying on expensive proprietary APIs.

\textbf{Limitations and Evaluation.} Due to high computational costs and the practical challenges of replicating existing agentic workflows, a direct empirical comparison with other contemporary systems remains for future work. Furthermore, while this study focuses on the synthesis framework itself, the downstream impact of the generated data on model fine-tuning has not yet been fully quantified. Systematically evaluating these performance gains and benchmarking TheoremForge against alternative architectures are essential next steps to validate the utility of our synthetic corpora.


\bibliography{refs}

@misc{chen2025seedprover15masteringundergraduatelevel,
      title={Seed-Prover 1.5: Mastering Undergraduate-Level Theorem Proving via Learning from Experience}, 
      author={Jiangjie Chen and Wenxiang Chen and Jiacheng Du and Jinyi Hu and Zhicheng Jiang and Allan Jie and Xiaoran Jin and Xing Jin and Chenggang Li and Wenlei Shi and Zhihong Wang and Mingxuan Wang and Chenrui Wei and Shufa Wei and Huajian Xin and Fan Yang and Weihao Gao and Zheng Yuan and Tianyang Zhan and Zeyu Zheng and Tianxi Zhou and Thomas Hanwen Zhu},
      year={2025},
      eprint={2512.17260},
      archivePrefix={arXiv},
      primaryClass={cs.CL},
}

@misc{chen2025seedproverdeepbroadreasoning,
      title={Seed-Prover: Deep and Broad Reasoning for Automated Theorem Proving}, 
      author={Luoxin Chen and Jinming Gu and Liankai Huang and Wenhao Huang and Zhicheng Jiang and Allan Jie and Xiaoran Jin and Xing Jin and Chenggang Li and Kaijing Ma and Cheng Ren and Jiawei Shen and Wenlei Shi and Tong Sun and He Sun and Jiahui Wang and Siran Wang and Zhihong Wang and Chenrui Wei and Shufa Wei and Yonghui Wu and Yuchen Wu and Yihang Xia and Huajian Xin and Fan Yang and Huaiyuan Ying and Hongyi Yuan and Zheng Yuan and Tianyang Zhan and Chi Zhang and Yue Zhang and Ge Zhang and Tianyun Zhao and Jianqiu Zhao and Yichi Zhou and Thomas Hanwen Zhu},
      year={2025},
      eprint={2507.23726},
      archivePrefix={arXiv},
      primaryClass={cs.AI},
}

@article{lin2025goedelproverv2,
  title={Goedel-Prover-V2: Scaling Formal Theorem Proving with Scaffolded Data Synthesis and Self-Correction},
  author={Lin, Yong and Tang, Shange and Lyu, Bohan and Yang, Ziran and Chung, Jui-Hui and Zhao, Haoyu and Jiang, Lai and Geng, Yihan and Ge, Jiawei and Sun, Jingruo and others},
  journal={arXiv preprint arXiv:2508.03613},
  year={2025}
}

@misc{wang2025ariaagentretrievaliterative,
      title={Aria: An Agent For Retrieval and Iterative Auto-Formalization via Dependency Graph}, 
      author={Hanyu Wang and Ruohan Xie and Yutong Wang and Guoxiong Gao and Xintao Yu and Bin Dong},
      year={2025},
      eprint={2510.04520},
      archivePrefix={arXiv},
      primaryClass={cs.AI},
}

@misc{varambally2025hilbertrecursivelybuildingformal,
      title={Hilbert: Recursively Building Formal Proofs with Informal Reasoning}, 
      author={Sumanth Varambally and Thomas Voice and Yanchao Sun and Zhifeng Chen and Rose Yu and Ke Ye},
      year={2025},
      eprint={2509.22819},
      archivePrefix={arXiv},
      primaryClass={cs.AI},
}

@misc{asher2025leanexploresearchenginelean,
      title={LeanExplore: A search engine for Lean 4 declarations}, 
      author={Justin Asher},
      year={2025},
      eprint={2506.11085},
      archivePrefix={arXiv},
      primaryClass={cs.SE},
}

@article{xin2025scaling,
  title={Scaling up Multi-Turn Off-Policy RL and Multi-Agent Tree Search for LLM Step-Provers},
  author={Xin, Ran and Zheng, Zeyu and Nie, Yanchen and Yuan, Kun and Xiao, Xia},
  journal={arXiv preprint arXiv:2509.06493},
  year={2025}
}

@misc{chen2025reform,
      title={ReForm: Reflective Autoformalization with Prospective Bounded Sequence Optimization}, 
      author={Guoxin Chen and Jing Wu and Xinjie Chen and Wayne Xin Zhao and Ruihua Song and Chengxi Li and Kai Fan and Dayiheng Liu and Minpeng Liao},
      year={2025},
      eprint={2510.24592},
      archivePrefix={arXiv},
      primaryClass={cs.CL},
}

@article{deepmath,
  title={DeepMath-103K: A Large-Scale, Challenging, Decontaminated, and  Verifiable Mathematical Dataset for Advancing Reasoning},
  author={He, Zhiwei and Liang, Tian and Xu, Jiahao and Liu, Qiuzhi and Chen, Xingyu and Wang, Yue and Song, Linfeng and Yu, Dian and Liang, Zhenwen and Wang, Wenxuan and Zhang, Zhuosheng and Wang, Rui and Tu, Zhaopeng and Mi, Haitao and Yu, Dong},
  year={2025},
  eprint={2504.11456},
  archivePrefix={arXiv},
  primaryClass={cs.CL},
}

@misc{zhang2025deeptheoremadvancingllmreasoning,
      title={DeepTheorem: Advancing LLM Reasoning for Theorem Proving Through Natural Language and Reinforcement Learning}, 
      author={Ziyin Zhang and Jiahao Xu and Zhiwei He and Tian Liang and Qiuzhi Liu and Yansi Li and Linfeng Song and Zhenwen Liang and Zhuosheng Zhang and Rui Wang and Zhaopeng Tu and Haitao Mi and Dong Yu},
      year={2025},
      eprint={2505.23754},
      archivePrefix={arXiv},
      primaryClass={cs.CL},
}

@inproceedings{wang2025dyflow,
title={DyFlow: Dynamic Workflow Framework for Agentic Reasoning},
author={Yanbo Wang and Zixiang Xu and Yue Huang and Xiangqi Wang and Zirui Song and Lang Gao and Chenxi Wang and Xiangru Tang and Yue Zhao and Arman Cohan and Xiangliang Zhang and Xiuying Chen},
booktitle={The Thirty-ninth Annual Conference on Neural Information Processing Systems},
year={2025},
}

@inproceedings{zhang2025aflow,
title={{AF}low: Automating Agentic Workflow Generation},
author={Jiayi Zhang and Jinyu Xiang and Zhaoyang Yu and Fengwei Teng and Xiong-Hui Chen and Jiaqi Chen and Mingchen Zhuge and Xin Cheng and Sirui Hong and Jinlin Wang and Bingnan Zheng and Bang Liu and Yuyu Luo and Chenglin Wu},
booktitle={The Thirteenth International Conference on Learning Representations},
year={2025},
}

@article{singh2505agentic,
  title={Agentic reasoning and tool integration for llms via reinforcement learning, 2025},
  author={Singh, Joykirat and Magazine, Raghav and Pandya, Yash and Nambi, Akshay},
  volume={5}
}

@article{nezhad2025symcode,
  title={SymCode: A Neurosymbolic Approach to Mathematical Reasoning via Verifiable Code Generation},
  author={Nezhad, Sina Bagheri and Li, Yao and Agrawal, Ameeta},
  journal={arXiv preprint arXiv:2510.25975},
  year={2025}
}

@inproceedings{ni-etal-2025-tree,
    title = "Tree-of-Code: A Self-Growing Tree Framework for End-to-End Code Generation and Execution in Complex Tasks",
    author = "Ni, Ziyi  and
      Li, Yifan  and
      Yang, Ning  and
      Shen, Dou  and
      Lyu, Pin  and
      Dong, Daxiang",
    editor = "Che, Wanxiang  and
      Nabende, Joyce  and
      Shutova, Ekaterina  and
      Pilehvar, Mohammad Taher",
    booktitle = "Findings of the Association for Computational Linguistics: ACL 2025",
    month = jul,
    year = "2025",
    address = "Vienna, Austria",
    publisher = "Association for Computational Linguistics",
    doi = "10.18653/v1/2025.findings-acl.509",
    pages = "9804--9819",
    ISBN = "979-8-89176-256-5"
}

@article{yuksekgonul2025optimizing,
  title={Optimizing generative AI by backpropagating language model feedback},
  author={Yuksekgonul, Mert and Bianchi, Federico and Boen, Joseph and Liu, Sheng and Lu, Pan and Huang, Zhi and Guestrin, Carlos and Zou, James},
  journal={Nature},
  volume={639},
  pages={609--616},
  year={2025},
}

@article{wang2024theoremllama,
  title={Theoremllama: Transforming general-purpose llms into lean4 experts},
  author={Wang, Ruida and Zhang, Jipeng and Jia, Yizhen and Pan, Rui and Diao, Shizhe and Pi, Renjie and Zhang, Tong},
  journal={arXiv preprint arXiv:2407.03203},
  year={2024}
}

@inproceedings{yao-etal-2025-fans,
    title = "{FANS}: Formal Answer Selection for {LLM} Natural Language Math Reasoning Using Lean4",
    author = "Yao, Jiarui  and
      Wang, Ruida  and
      Zhang, Tong",
    editor = "Christodoulopoulos, Christos  and
      Chakraborty, Tanmoy  and
      Rose, Carolyn  and
      Peng, Violet",
    booktitle = "Proceedings of the 2025 Conference on Empirical Methods in Natural Language Processing",
    month = nov,
    year = "2025",
    address = "Suzhou, China",
    publisher = "Association for Computational Linguistics",
    doi = "10.18653/v1/2025.emnlp-main.158",
    pages = "3181--3200",
    ISBN = "979-8-89176-332-6"
}

@article{wang2025evoagentx,
  title={EvoAgentX: An Automated Framework for Evolving Agentic Workflows},
  author={Wang, Yingxu and Liu, Siwei and Fang, Jinyuan and Meng, Zaiqiao},
  journal={arXiv preprint arXiv:2507.03616},
  year={2025}
}

@inproceedings{xu-etal-2025-aristotle,
    title = "Aristotle: Mastering Logical Reasoning with A Logic-Complete Decompose-Search-Resolve Framework",
    author = "Xu, Jundong  and
      Fei, Hao  and
      Luo, Meng  and
      Liu, Qian  and
      Pan, Liangming  and
      Wang, William Yang  and
      Nakov, Preslav  and
      Lee, Mong-Li  and
      Hsu, Wynne",
    editor = "Che, Wanxiang  and
      Nabende, Joyce  and
      Shutova, Ekaterina  and
      Pilehvar, Mohammad Taher",
    booktitle = "Proceedings of the 63rd Annual Meeting of the Association for Computational Linguistics (Volume 1: Long Papers)",
    month = jul,
    year = "2025",
    address = "Vienna, Austria",
    publisher = "Association for Computational Linguistics",
    doi = "10.18653/v1/2025.acl-long.153",
    pages = "3052--3075",
    ISBN = "979-8-89176-251-0"
}

@article{hubert2025olympiad,
  title={Olympiad-level formal mathematical reasoning with reinforcement learning},
  author={Hubert, Thomas and Mehta, Rishi and Sartran, Laurent and Horv{\'a}th, Mikl{\'o}s Z and {\v{Z}}u{\v{z}}i{\'c}, Goran and Wieser, Eric and Huang, Aja and Schrittwieser, Julian and Schroecker, Yannick and Masoom, Hussain and others},
  journal={Nature},
  pages={1--3},
  year={2025},
  publisher={Nature Publishing Group UK London}
}

@inproceedings{
ospanov2025apollo,
title={{APOLLO}: Automated {LLM} and Lean Collaboration for Advanced Formal Reasoning},
author={Azim Ospanov and Farzan Farnia and Roozbeh Yousefzadeh},
booktitle={The Thirty-ninth Annual Conference on Neural Information Processing Systems},
year={2025},
}

@inproceedings{
baba2025prover,
title={Prover Agent: An Agent-based Framework for Formal Mathematical Proofs},
author={Kaito Baba and Chaoran Liu and Shuhei Kurita and Akiyoshi Sannai},
booktitle={2nd AI for Math Workshop @ ICML 2025},
year={2025},
}

@misc{hu2025hybridproveraugmentingtheoremproving,
      title={HybridProver: Augmenting Theorem Proving with LLM-Driven Proof Synthesis and Refinement}, 
      author={Jilin Hu and Jianyu Zhang and Yongwang Zhao and Talia Ringer},
      year={2025},
      eprint={2505.15740},
      archivePrefix={arXiv},
      primaryClass={cs.FL},
}

@misc{jana2025proofbridgeautoformalizationnaturallanguage,
      title={ProofBridge: Auto-Formalization of Natural Language Proofs in Lean via Joint Embeddings}, 
      author={Prithwish Jana and Kaan Kale and Ahmet Ege Tanriverdi and Cruise Song and Sriram Vishwanath and Vijay Ganesh},
      year={2025},
      eprint={2510.15681},
      archivePrefix={arXiv},
      primaryClass={cs.LO},
}

@misc{
lu2025processdriven,
title={Process-Driven Autoformalization in Lean 4},
author={Jianqiao Lu and Yingjia Wan and Zhengying Liu and Yinya Huang and Jing Xiong and Liu Chengwu and Jianhao Shen and Hui Jin and Jipeng Zhang and Haiming Wang and Zhicheng Yang and Jing Tang and Zhijiang Guo},
year={2025},
}

@inproceedings{
gao2025herald,
title={Herald: A Natural Language Annotated Lean 4 Dataset},
author={Guoxiong Gao and Yutong Wang and Jiedong Jiang and Qi Gao and Zihan Qin and Tianyi Xu and Bin Dong},
booktitle={The Thirteenth International Conference on Learning Representations},
year={2025},
}

@article{kimina_prover_2025,
    title = {Kimina-Prover Preview: Towards Large Formal Reasoning Models with Reinforcement Learning},
    author = {Wang, Haiming and Unsal, Mert and Lin, Xiaohan and Baksys, Mantas and Liu, Junqi and Santos, Marco Dos and Sung, Flood and Vinyes, Marina and Ying, Zhenzhe and Zhu, Zekai and Lu, Jianqiao and Saxcé, Hugues de and Bailey, Bolton and Song, Chendong and Xiao, Chenjun and Zhang, Dehao and Zhang, Ebony and Pu, Frederick and Zhu, Han and Liu, Jiawei and Bayer, Jonas and Michel, Julien and Yu, Longhui and Dreyfus-Schmidt, Léo and Tunstall, Lewis and Pagani, Luigi and Machado, Moreira and Bourigault, Pauline and Wang, Ran and Polu, Stanislas and Barroyer, Thibaut and Li, Wen-Ding and Niu, Yazhe and Fleureau, Yann and Hu, Yangyang and Yu, Zhouliang and Wang, Zihan and Yang, Zhilin and Liu, Zhengying and Li, Jia},
    year = {2025},
}

@inproceedings{gao-etal-2024-semantic-search,
    title = "A Semantic Search Engine for Mathlib4",
    author = "Gao, Guoxiong  and
      Ju, Haocheng  and
      Jiang, Jiedong  and
      Qin, Zihan  and
      Dong, Bin",
    editor = "Al-Onaizan, Yaser  and
      Bansal, Mohit  and
      Chen, Yun-Nung",
    booktitle = "Findings of the Association for Computational Linguistics: EMNLP 2024",
    month = nov,
    year = "2024",
    address = "Miami, Florida, USA",
    publisher = "Association for Computational Linguistics",
    doi = "10.18653/v1/2024.findings-emnlp.470",
    pages = "8001--8013"
}

@inproceedings{
liu2025rethinking,
title={Rethinking and Improving Autoformalization: Towards a Faithful Metric and a Dependency Retrieval-based Approach},
author={Qi Liu and Xinhao Zheng and Xudong Lu and Qinxiang Cao and Junchi Yan},
booktitle={The Thirteenth International Conference on Learning Representations},
year={2025},
}

@inproceedings{
jiang2023draft,
title={Draft, Sketch, and Prove: Guiding Formal Theorem Provers with Informal Proofs},
author={Albert Qiaochu Jiang and Sean Welleck and Jin Peng Zhou and Timothee Lacroix and Jiacheng Liu and Wenda Li and Mateja Jamnik and Guillaume Lample and Yuhuai Wu},
booktitle={The Eleventh International Conference on Learning Representations },
year={2023},
}

@inproceedings{
xin2025deepseekproverv,
title={DeepSeek-Prover-V1.5: Harnessing Proof Assistant Feedback for Reinforcement Learning and Monte-Carlo Tree Search},
author={Huajian Xin and Z.Z. Ren and Junxiao Song and Zhihong Shao and Wanjia Zhao and Haocheng Wang and Bo Liu and Liyue Zhang and Xuan Lu and Qiushi Du and Wenjun Gao and Haowei Zhang and Qihao Zhu and Dejian Yang and Zhibin Gou and Z.F. Wu and Fuli Luo and Chong Ruan},
booktitle={The Thirteenth International Conference on Learning Representations},
year={2025},
}

@misc{ren2025deepseekproverv2advancingformalmathematical,
      title={DeepSeek-Prover-V2: Advancing Formal Mathematical Reasoning via Reinforcement Learning for Subgoal Decomposition}, 
      author={Z. Z. Ren and Zhihong Shao and Junxiao Song and Huajian Xin and Haocheng Wang and Wanjia Zhao and Liyue Zhang and Zhe Fu and Qihao Zhu and Dejian Yang and Z. F. Wu and Zhibin Gou and Shirong Ma and Hongxuan Tang and Yuxuan Liu and Wenjun Gao and Daya Guo and Chong Ruan},
      year={2025},
      eprint={2504.21801},
      archivePrefix={arXiv},
      primaryClass={cs.CL},
}

@inproceedings{yang2023leandojo,
  title={{LeanDojo}: Theorem Proving with Retrieval-Augmented Language Models},
  author={Yang, Kaiyu and Swope, Aidan and Gu, Alex and Chalamala, Rahul and Song, Peiyang and Yu, Shixing and Godil, Saad and Prenger, Ryan and Anandkumar, Anima},
  booktitle={Neural Information Processing Systems (NeurIPS)},
  year={2023}
}

@article{jiang2025fate,
  title={Fate: A formal benchmark series for frontier algebra of multiple difficulty levels},
  author={Jiang, Jiedong and He, Wanyi and Wang, Yuefeng and Gao, Guoxiong and Hu, Yongle and Wang, Jingting and Guan, Nailing and Wu, Peihao and Dai, Chunbo and Xiao, Liang and others},
  journal={arXiv preprint arXiv:2511.02872},
  year={2025}
}

@inproceedings{icml2025_supernet,
title={Multi-agent Architecture Search via Agentic Supernet},
author={Guibin Zhang and Luyang Niu and Junfeng Fang and Kun Wang and LEI BAI and Xiang Wang},
booktitle={Forty-second International Conference on Machine Learning},
year={2025},
}

@inproceedings{iclr2025_adaptive,
title={Adaptive Self-improvement {LLM} Agentic System for {ML} Library Development},
author={Genghan Zhang and Weixin Liang and Olivia Hsu and Kunle Olukotun},
booktitle={ICLR 2025 Third Workshop on Deep Learning for Code},
year={2025},
}

@inproceedings{icml2025_collaborative,
  title={Collaborative Mean Estimation Among Heterogeneous Strategic Agents: Individual Rationality, Fairness, and Truthful Contribution},
  author={Clinton, Alex and Chen, Yiding and Zhu, Jerry and Kandasamy, Kirthevasan},
  booktitle={Forty-second International Conference on Machine Learning}
}

@inproceedings{icml2025_trustworthy,
title={Position: Trustworthy {AI} Agents Require the Integration of Large Language Models and Formal Methods},
author={Yedi Zhang and Yufan Cai and Xinyue Zuo and Xiaokun Luan and Kailong Wang and Zhe Hou and Yifan Zhang and Zhiyuan Wei and Meng Sun and Jun Sun and Jing Sun and Jin Song Dong},
booktitle={Forty-second International Conference on Machine Learning Position Paper Track},
year={2025},
}

@article{hu2025qualityflow,
  title={Qualityflow: An agentic workflow for program synthesis controlled by llm quality checks},
  author={Hu, Yaojie and Zhou, Qiang and Chen, Qihong and Li, Xiaopeng and Liu, Linbo and Zhang, Dejiao and Kachroo, Amit and Oz, Talha and Tripp, Omer},
  journal={arXiv preprint arXiv:2501.17167},
  year={2025}
}

@inproceedings{kozyrev2025rocqstar,
  title={RocqStar: Leveraging Similarity-driven Retrieval and Agentic Systems for Rocq generation},
  author={Kozyrev, Andrei and Khramov, Nikita and Solovev, Gleb and Podkopaev, Anton},
  booktitle={NeurIPS 2025 Fourth Workshop on Deep Learning for Code}
}

@inproceedings{wang2025qdtsynth,
  title={QDTSynth: Quality-Driven Formal Theorem Synthesis for Enhancing Proving Performance of LLMs},
  author={Wang, Lei and Zuo, Ruobing and He, Gaolei and Wang, Jianlin and Yang, Zhengfeng},
  booktitle={Proceedings of the 63rd Annual Meeting of the Association for Computational Linguistics (Volume 1: Long Papers)},
  pages={14683--14698},
  year={2025}
}


\newpage
\appendix
\section*{Appendix}
\section{Implementation and Configuration}\label{appendix:config}

In this section, we present a detailed demonstration of the implementation of TheoremForge workflow and the inference configuration. Some of the prompts are taken from Hilbert~\citep{varambally2025hilbertrecursivelybuildingformal}, which will not be presented here. These include prompts for search query generation, premise selection, proof sketch generation, iterative proof refinement, subgoal extraction and proof assembly. You can find the details in our code. Our code is open-sourced\footnote{\url{https://github.com/timechess/TheoremForge.git}}.

\subsection{Details on Closed-Source Models and Reproducibility}

All closed-source large language models used in our experiments were accessed via commercial API services. Except for Qwen-Max, all models were invoked through the CloseAI aggregation platform\footnote{\url{https://www.closeai-asia.com/}}. Qwen-Max was accessed through the Alibaba Cloud Large Model Platform \footnote{\url{https://bailian.console.aliyun.com/}}. GPT-5.2, Claude Sonnet 4.5, DeepSeek models, and Qwen-Max were called using an OpenAI-compatible SDK, while Gemini models were accessed via the official Google GenAI SDK.

For the concrete model identifiers used in our experiments, see Table~\ref{tab:model_pricing}. All experiments were conducted in December 2025 using the latest available model versions at the time.

For inference configuration, the sampling temperature was set to 0 for all non-Gemini models. For Gemini models, we followed the provider’s recommendation and set the temperature to 1.0. All other inference parameters were left at their platform-default values.

All evaluated models except for DeepSeek-V3.2 are proprietary and closed-source, and explicit version hashes were not exposed through the APIs. Moreover, these models may undergo unannounced updates by the service providers. We therefore report the exact model identifiers, access platforms, SDKs, inference parameters, and the evaluation time window to maximize reproducibility. While exact bit-level reproducibility cannot be guaranteed, these details sufficiently constrain the experimental setting to enable meaningful comparison under comparable API

The local expert models are served on a 4xA100 server. The prices are calculated using Table~\ref{tab:model_pricing}.

\begin{table}[ht]
\centering
\caption{Identifier and pricing of evaluated language models (\$/M token).}
\label{tab:model_pricing}
\begin{tabular}{llcc}
\toprule
\textbf{Model} & \textbf{Identifier}& \textbf{Input} & \textbf{Output} \\
\midrule
Gemini-3-Pro              & \texttt{gemini-3-pro-preview} & 2.00 & 12.00 \\
Gemini-3-Flash            & \texttt{gemini-3-flash-preview} & 0.50 & 3.00  \\
GPT-5.2                   & \texttt{gpt-5.2} & 1.75 & 14.00 \\
DeepSeek-V3.2             & \texttt{deepseek-chat} & 0.55 & 1.70  \\
DeepSeek-V3.2-Thinking    & \texttt{deepseek-reasoner} & 0.55 & 1.70  \\
Qwen-Max                  & \texttt{qwen-max}& 0.46 & 1.84  \\
Claude-Sonnet-4.5         & \texttt{claude-sonnet-4-5} & 3.00 & 15.00 \\
\bottomrule
\end{tabular}
\end{table}

\subsection{Statement Formalization Workflow}
\textbf{Statement Normalization}. Input statements vary in format and often contain implicit premises or lack explicit answers. This step preprocesses and normalizes the inputs to facilitate accurate downstream translation. 

\textbf{Definition Retrieval}. Based on the normalized statement, we prompt the LLM to generate up to $k_{\text{query}}=5$ queries to search for relevant definitions. We utilize the open-source search engine LeanExplore~\citep{asher2025leanexploresearchenginelean} to retrieve the top-5 most similar constants from Mathlib for each query. Subsequently, we prompt the LLM selects the most relevant constants from these results.

\textbf{Expert Sampling}. Given the normalized statement and the selected constants, we employ an expert model to generate $k_{\text{formalizer}}=4$ candidates and perform syntax checks using the compiler. If no valid code is generated, we prompt the LLM to attempt a fix based on the compiler's error messages. If not otherwise mentioned, we only give one chance for correction to each failed code in this section. 

\textbf{Semantic Check}. All candidates that pass the syntax check undergo semantic verification, where the LLM assesses their consistency with the original input semantics. The final justification will be a binary signal, which is ``ALIGNED'' or ``NOT\_ALIGNED''. If the LLM identifies an inconsistency, it is prompted to provide a correction. If the corrected code compiles successfully, it is deemed to have passed the semantic check.

\textbf{Formalization Selection}. Following the filtering steps, we instruct the LLM to select the optimal candidate as the final output based on criteria such as semantic consistency and estimated proof difficulty. The model is also required to provide a rationale for its selection.

\subsection{Proof Generation}
\textbf{Expert Sampling}. We invoke an expert prover model to generate $k_{\text{prover}}=4$ candidate proofs. If all candidates fail, we prompt the LLM to correct the proofs. We observe that most open-source prover models are not trained in a retrieval-augmented style. Consequently, providing auxiliary theorems in the context offers negligible performance benefits at this stage.

\textbf{Theorem Retrieval}. This phase mirrors the definition retrieval process used in the statement formalization workflow.

\textbf{Informal Proof Generation}. Given the formal statement and the retrieval results, we prompt the LLM to generate a step-by-step informal proof to serve as a reference for the subsequent decomposition.

\textbf{Subgoal Decomposition}. Using the formal statement, retrieval results, and the informal proof, we prompt the LLM to generate a formal proof sketch. This sketch includes several intermediate steps (using \texttt{have} statements) and the logic connecting them. We then leverage the LLM to extract these \texttt{have} statements into independent lemmas. We rely on LLM-based extraction rather than the \texttt{extract\_goals} tactic, as the latter has proven unreliable in certain contexts. If the proof sketch fails the compiler check, the LLM is prompted to attempt a fix.

\textbf{Subgoal Solving}. The extracted subgoals undergo the expert sampling stage. We do not perform new retrieval for these subgoals. Instead, we reuse the retrieval results from the root problem. Still, we first sample from the expert prover with $k_\text{prover}=4$ attempts and try to correct the failed proofs if all fail. If the expert prover cannot provide valid proofs with one chance to correct, we prompt the general-purpose LLM to generate an informal proof for this subgoal. Then the LLM is required to generate the formal proof according to the informal proof, along with useful theorems retrieved in the stage ahead. We set $k_{\text{refine}}=2$ to allow at most 2 attempts to fix failed proof.

\textbf{Proof Assembly}. Once all subgoals are solved, we prompt the LLM to synthesize the final proof by assembling the sub-proofs according to the original sketch. If the assembly fails, we give the LLM one chance to fix. If one subgoal fails to be proved, then the other subgoals in the queue will be cancelled, serving as an \textit{Early Stop} policy to save budget.

\subsection{Semantic Verification}\label{appendix:semantic}

Here we provide a detailed specification of the semantic verification process.

Let $\mathcal{P} = \{p_1, \dots, p_N\}$ denote a set of mathematical problems expressed in natural language.
For each problem $p_i$, a proof generation model produces a formal statement
$\hat{y}_i$.
We only perform semantic verifications for formal statements that succeed in generating proofs. Assuming there are $M \le N$ problems satisifying the condition, denoted as $\{p'_1, \dots, p'_{M}\}$.

\paragraph{Verifier Models and Identities.}
We consider a set of $K=7$ LLM-based semantic verifier models
\[
\mathcal{E} = \{E_1, E_2, \dots, E_7\}.
\]
Each verifier $E_k$ is associated with a verifier identity $\mathrm{id}(E_k)$.
In particular, different variants of the same underlying model (DeepSeek-V3.2 with and without thinking) are treated as sharing the same identity.

Each verifier takes as input a pair $(p'_i, \hat{y}_i)$, where $p'_i$ is the original
natural-language proposition and $\hat{y}'_i$ is the generated formal output, and outputs a binary judgment
\[
E_k(p'_i, \hat{y}'_i) \in \{0,1\},
\]
with $1$ indicating that $\hat{y}'_i$ is judged to be semantically consistent with the
meaning of $p'_i$, and $0$ otherwise.

\paragraph{Exclusion of Generator from Voting.}
Let $G_i$ denote the generator model that produces $\hat{y}'_i$.
To avoid self-evaluation bias, all verifier models whose identities coincide with that of the generator are excluded from voting.
Formally, the effective verifier set for $(p'_i, \hat{y}'_i)$ is defined as
\[
\mathcal{E}_i =
\{E_k \in \mathcal{E} \mid \mathrm{id}(E_k) \neq \mathrm{id}(G_i)\}.
\]

For Baseline, we do not exclude any model.

\paragraph{Majority-Based Verification Rule.}
Let $|\mathcal{E}_i|$ denote the number of effective verifiers for problem $p_i$. We define the aggregated semantic verification decision $V(p_i, \hat{y}_i)$ using majority voting over $\mathcal{E}_i$:
\[
V(p'_i, \hat{y}'_i) =
\begin{cases}
1, & \text{if } \sum\limits_{E_k \in \mathcal{E}_i} E_k(p'_i, \hat{y}'_i)
      \ge \left\lceil \dfrac{|\mathcal{E}_i|}{2} \right\rceil, \\
0, & \text{otherwise}.
\end{cases}
\]


\paragraph{Verified Rate.}
The \emph{Verified Rate (VR)} is defined as the fraction of problems whose generated outputs are semantically verified under the above aggregation rule while having a valid proof:
\[
\mathrm{VR} = \frac{1}{N} \sum_{i=1}^{M} V(p'_i, \hat{y}'_i).
\]

\paragraph{Conservative and Lenient Variants.}
For completeness, we also define two alternative verification criteria based on the effective verifier set $\mathcal{E}_i$:
\begin{align*}
&V_{\mathrm{strict}}(p_i, \hat{y}_i)
= \mathbb{I}\!\left[\sum_{E_k \in \mathcal{E}_i} E_k(p_i, \hat{y}_i)
= |\mathcal{E}_i| \right], \\
&V_{\mathrm{lenient}}(p_i, \hat{y}_i)
= \mathbb{I}\!\left[\sum_{E_k \in \mathcal{E}_i} E_k(p_i, \hat{y}_i)
\ge 1 \right],
\end{align*}
which provide conservative and optimistic bounds on the reported verified rate, respectively.

\begin{figure*}[h]
    \centering
    \begin{subfigure}[b]{0.62\textwidth}
        \centering
        \includegraphics[height=5.5cm, keepaspectratio]{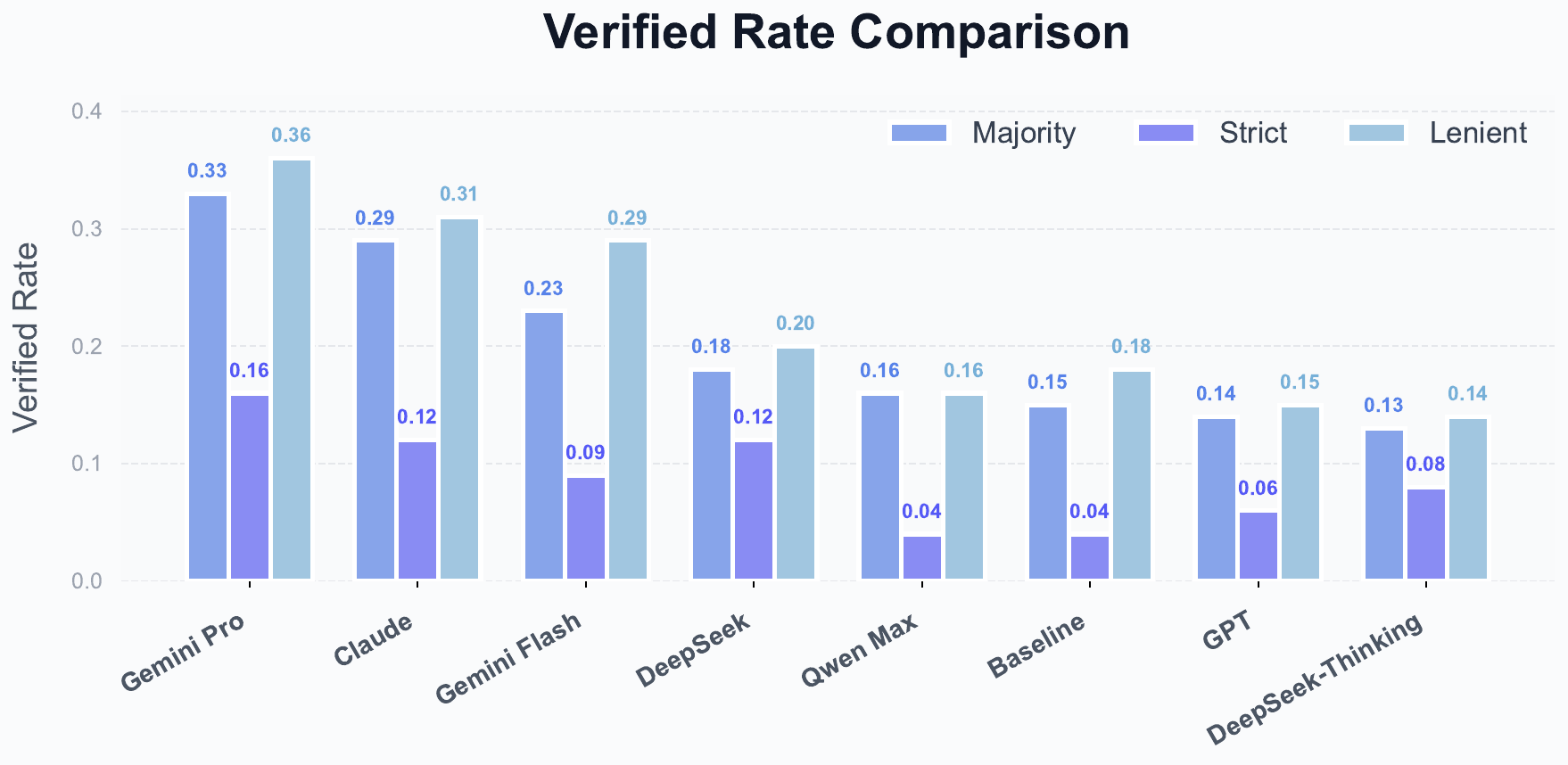} 
        \caption{}
        \label{fig:vr_bar_appendix}
    \end{subfigure}
    \hfill
    \begin{subfigure}[b]{0.36\textwidth}
        \centering
        \includegraphics[height=5.5cm, keepaspectratio]{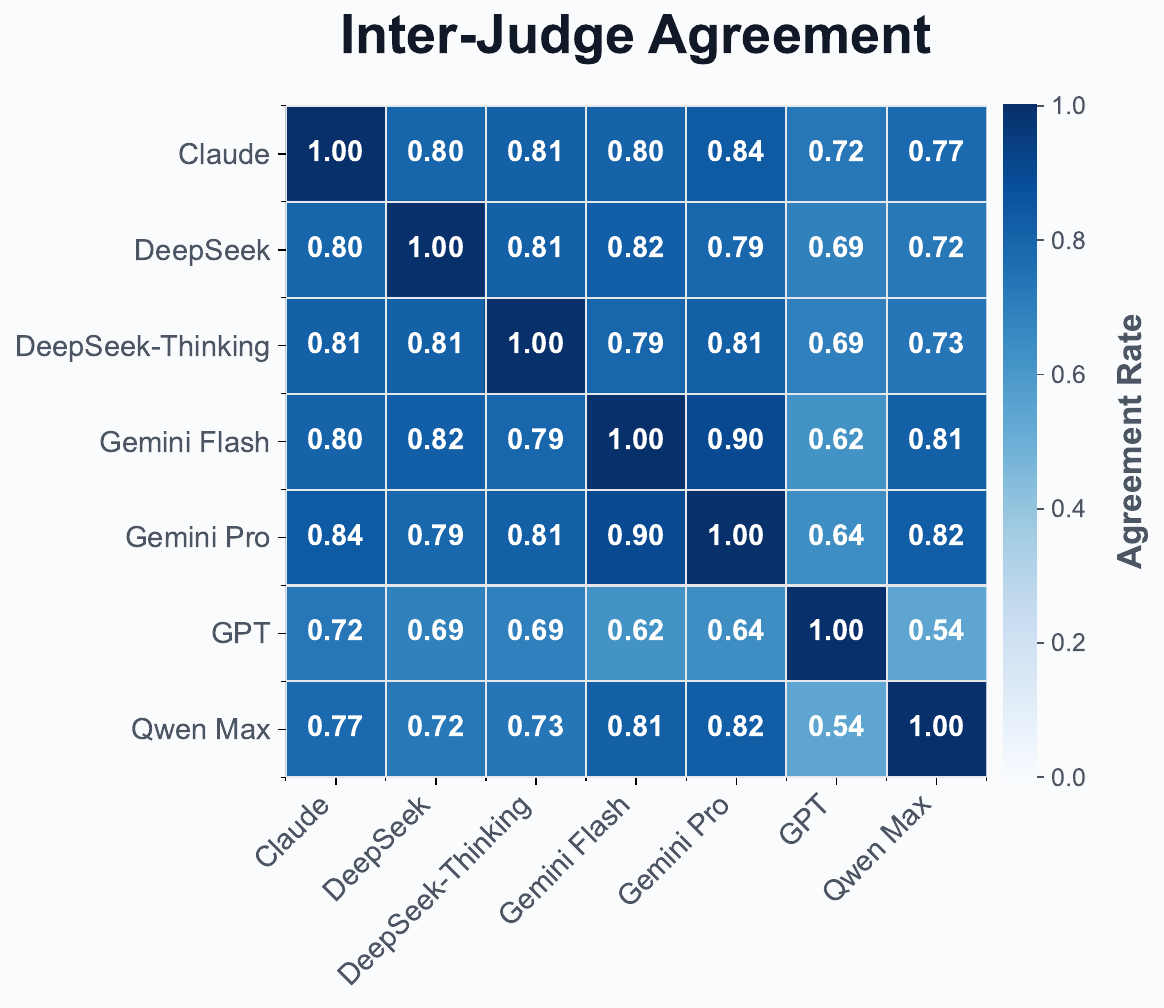} 
        \caption{}
        \label{fig:inter_judge_appendix}
    \end{subfigure}
    
    \caption{\textbf{Quantitative analysis of the semantic verification process.} 
    (a) Comparison of verified rate (VR) on the small-scale benchmark under Majority, Strict, and Lenient aggregation rules. 
    (b) Heatmap of pairwise agreement rates between the seven LLM verifiers. 
    Higher values (darker blue) indicate stronger consensus between models. 
    The results are calculated using verification data from large-scale benchmark (including TheoremForge and Baseline).}
    \label{fig:semantic_judge}
\end{figure*}

\paragraph{Results on Semantic Verification.}
The results for verified rates (VR) across different models and aggregation rules are summarized in Table~\ref{tab:semantic_vr} and Figure~\ref{fig:vr_bar_appendix}. 
Among all evaluated models on the small-scale benchmark, \textbf{Gemini-3-Pro (low)} and \textbf{Claude-Sonnet-4.5} demonstrate the highest formalization quality, achieving 33.0\% and 29.0\% Majority VR, respectively. 
Notably, we observe a substantial gap between Strict VR and Lenient VR; for instance, Gemini-3-Pro drops from 36.0\% (Lenient) to 16.0\% (Strict), suggesting that while most models can produce formalizations recognized as correct by at least one verifier, achieving universal consensus across all LLM judges remains a high bar. 
On the large-scale benchmark, \textbf{TheoremForge} significantly outperforms the baseline (12.6\% vs. 8.6\% Majority VR), validating the efficacy of our proposed framework in maintaining semantic fidelity during large-scale formalization tasks.

\begin{table}[h]
\centering
\captionsetup{width=0.53\columnwidth}

\caption{Semantic verified rates (VR) under different aggregation rules on
small-scale and large-scale benchmarks.}
\label{tab:semantic_vr}
\small
\setlength{\tabcolsep}{6pt}
\begin{tabular}{l
                S[table-format=2.1]
                S[table-format=2.1]
                S[table-format=2.1]}
\toprule
Model / Method & {Majority VR} & {Strict VR} & {Lenient VR} \\
\midrule
\multicolumn{4}{l}{\textbf{Small-Scale Benchmark}} \\
\addlinespace[2pt]
GPT-5.2                 & 14.0 & 6.0  & 15.0 \\
Claude-Sonnet-4.5       & 29.0 & 12.0 & 31.0 \\
Gemini-3-Flash (low)    & 23.0 & 9.0  & 29.0 \\
Gemini-3-Pro (low)      & 33.0 & 16.0 & 36.0 \\
DeepSeek-V3.2-Thinking  & 13.0 & 8.0  & 14.0 \\
DeepSeek-V3.2           & 18.0 & 12.0 & 20.0 \\
Qwen-Max                & 16.0 & 4.0  & 16.0 \\
Baseline                & 15.0 & 4.0  & 18.0 \\
\addlinespace[4pt]
\multicolumn{4}{l}{\textbf{Large-Scale Benchmark}} \\
\addlinespace[2pt]
TheoremForge             & 12.6 & 6.2 & 14.2 \\
Baseline                 & 8.6  & 4.2 & 11.9 \\
\bottomrule
\end{tabular}
\end{table}

\paragraph{Inter-Judge Reliability.}
To evaluate the consistency of the LLM-based ensemble, we analyze the inter-judge agreement between verifier models. The agreement rate $A(E_a, E_b)$ between two verifiers $E_a, E_b \in \mathcal{E}$ is defined as the fraction of instances where they provide identical judgments, calculated over the set of problems where both models are eligible to vote. Formally:
\begin{equation*}
A(E_a, E_b) = \frac{\sum_{i=1}^{M} \mathbb{I}[E_a \in \mathcal{E}_i \land E_b \in \mathcal{E}_i \land E_a(p'_i, \hat{y}'_i) = E_b(p'_i, \hat{y}'_i)]}{\sum_{i=1}^{M} \mathbb{I}[E_a \in \mathcal{E}_i \land E_b \in \mathcal{E}_i]}
\end{equation*}
where $\mathbb{I}[\cdot]$ is the indicator function. Note that $A(E_a, E_a) = 1.0$ by definition.

As illustrated in Figure~\ref{fig:inter_judge_appendix}, the majority of verifier pairs exhibit robust consensus, with agreement rates typically exceeding $0.75$. The Gemini family shows the highest internal consistency, with an agreement rate of $0.90$ between Gemini Flash and Gemini Pro. Claude also demonstrates strong alignment with both the Gemini and DeepSeek series, maintaining scores between $0.80$ and $0.84$. 

However, GPT stands out as a relative outlier, exhibiting a noticeably lower agreement with all other verifiers in the ensemble. Specifically, its agreement rates with other frontier models range from a minimum of $0.54$ (with Qwen Max) to a maximum of only $0.72$ (with Claude). This systematic divergence suggests that GPT may employ a fundamentally different or more idiosyncratic internal rubric for assessing semantic equivalence compared to its peers. Such results underscore the importance of our ensemble-based majority voting, as it prevents the specific biases or divergent judgment patterns of a single model like GPT from disproportionately affecting the final verified rate.

\subsection{Prompts}
\begin{resultbox}{Prompt for Statement Normalization}
\begin{Verbatim}[breaklines=true, breakanywhere=true, fontsize=\footnotesize]
You are helping to prepare an informal mathematical statement for formalization into Lean.
The input statement may be a math word problem, a question, or an incomplete statement that needs to be transformed into a clear mathematical proposition.

Normalization Process:
1. Convert questions into declarative statements:
   - "What is the value of x?" -> "The value of x is [answer]" or "There exists a unique x such that..."
   - "Is f continuous?" -> "f is continuous" or "f is not continuous"
   - "Can we prove that...?" -> "It holds that..." or "The statement ... is true"

2. Transform word problems into formal propositions:
   - Extract the mathematical content from the narrative
   - Identify all variables, constraints, and relationships
   - State what needs to be proven or computed as a clear proposition

3. Make implicit statements explicit:
   - Add missing quantifiers (for all, there exists)
   - Specify domains and ranges of variables
   - If the constraints can be inferred or proved from the problem, do not add them to the normalized statement

4. Structure the proposition clearly:
   - State assumptions/hypotheses first
   - Then state the conclusion or claim
   - Use standard mathematical language and notation

Guidelines:
- If the statement is already a clear proposition, you may keep it as is
- If it's a question without a known answer, solve the problem and state the answer along with the constraints as a proposition
- If it's a question with an answer provided, state the answer along with the constraints as a proposition
- Maintain the mathematical content and intent of the original statement
- Use precise mathematical language suitable for formalization
- Check whether the normalized statement is semantically aligned with the informal statement
- DO NOT only output the answer to the problem, you should state the answer along with the constraints as a proposition.

Input Statement:

{{ informal_statement }}

Output Format:
Provide the normalized statement in <normalized> tags.

Please reason step by step before providing the normalized statement.
\end{Verbatim}
\end{resultbox}

\begin{resultbox}{Prompt for Semantic Check}
\begin{Verbatim}[breaklines=true, breakanywhere=true, fontsize=\footnotesize]
You are an expert in mathematical formalization and Lean 4. Your task is to verify that a formal Lean 4 statement accurately captures the mathematical meaning of the original informal statement.

Original Informal Statement:
{{ informal_statement }}

{% if normalized_statement %}
Normalized Statement:
{{ normalized_statement }}
{% endif %}

{% if useful_definitions %}
Useful Definitions:
{{ useful_definitions }}
{% endif %}

Generated Formal Statement (Lean 4):
```lean4
{{ formal_statement }}
```

Your Task:
Carefully analyze whether the formal Lean 4 statement is semantically aligned with the original informal statement{% if normalized_statement %} and the normalized statement{% endif %}. Note that the normalized statement may contain information (e.g. the answer to a question) that is not present in the informal statement. Check whether the normalized statement is aligned with the informal statement. Then check whether the formal statement is aligned with the normalized statement. If NOT_ALIGNED, try to fix the formal statement to be aligned with the informal statement.

Evaluation Criteria:
1. **Mathematical Equivalence**: Does the formal statement express the same mathematical claim as the normalized statement?
2. **Completeness**: Are all conditions, hypotheses, and conclusions from the normalized statement captured in the formal statement?
3. **Correctness**: Are the mathematical objects, operations, and properties correctly formalized?
4. **Type Accuracy**: Are the types and type classes appropriate for the mathematical concepts?
5. **Quantifiers**: Are the quantifiers used correctly and do they match the informal statement?
6. **No Extra Constraints**: Does the formal statement avoid introducing additional constraints not present in the normalized statement? Or does the normalized statement contain additional constraints that are not present in the informal statement?
7. **No Missing Constraints**: Does the formal statement include all constraints from the normalized statement?
8. **Provability**: Is the formal statement provable given the provided premises?

Analysis Process:
1. Identify the key mathematical concepts in the informal statement
2. Check how each concept is formalized in the Lean 4 code
3. Verify that the logical structure is preserved
4. Look for any discrepancies, missing elements, or extra constraints
5. Check whether the formal statement is provable given the provided premises

Output Format:
Provide your analysis in the following structure:

<analysis>
[Provide a detailed step-by-step analysis comparing the informal and formal statements]
</analysis>

<verdict>
[Output EXACTLY one of: "ALIGNED" or "NOT_ALIGNED"]
</verdict>

<fixed_formal_statement>
[Output the fixed formal statement inside ```lean4``` code blocks. You should only output ONE fixed formalization at the end.]
</fixed_formal_statement>
\end{Verbatim}
\end{resultbox}

\begin{resultbox}{Prompt for Formalization Selection}
\begin{Verbatim}[breaklines=true, breakanywhere=true, fontsize=\footnotesize]
You are an expert in Lean 4 theorem proving. Your task is to select the BEST formalization to prove from multiple semantically aligned formalizations of the same informal statement.

**Informal Statement:**
{{ informal_statement }}

**Available Formalizations:**
{% for idx, formalization in formalizations %}
**Formalization {{ idx + 1 }}:**
```lean4
{{ formalization }}
```
{% endfor %}

**Instructions:**
1. Analyze each formalization carefully, considering:
   - Structural complexity
   - How many hypotheses need to be handled
   - Whether the formalization uses complex type theory constructs
   - Whether it requires proving stronger/more general results
   - Availability of library theorems that might help
   - How natural the proof approach would be
   - Avoid using too general definitions that are hard to compute. (e.g. `Euclidean.dist` in real space or general `Field` instead of $\mathbb{Q}$.)
   - If the informal statement contains quantifiers, the formalization should also contain quantifiers.

2. The BEST formalization should be EASY to prove while maintaining the mathematical meaning of the informal statement.

3. Provide your analysis in the following format:
   - Wrap your reasoning in <analysis></analysis> tags
   - Specify the selected formalization number (1, 2, 3, etc.) in <selected>number</selected> tags

**Example Response Format:**
<analysis>
Your detailed reasoning about why you chose this formalization...
</analysis>
<selected>2</selected>
\end{Verbatim}
\end{resultbox}

\begin{resultbox}{Prompt for Semantic Verification}
\begin{Verbatim}[breaklines=true, breakanywhere=true, fontsize=\footnotesize]
Role: Lean & Formal Verification Expert

Input:

- Mathematical_Text: A math problem and its answer (no proof).
- Lean4Code: A Lean 4 theorem statement formalizing the problem. Proof is intentionally omitted (e.g., sorry).

Goal:
Determine if the Lean theorem statement is an exact and faithful formalization of the mathematical problem.  
**Do not evaluate or consider the answer or the proof. Your sole task is to verify the correctness of the formalization.**

Evaluation Stages (All required):

1. Math Assertion Analysis  
   Identify all structurally and semantically relevant components of the mathematical problem, including variables, types, quantifiers, constraints, logic structure, conclusion, and so on. The analysis should be based on the actual content of the text.

2. Lean Statement Analysis (ignore proof part)  
   Extract all structurally and semantically relevant components from the Lean statement, including variables, types, conditions, quantifiers, constraints, the final claim, and so on. The analysis should reflect the actual content present in the Lean code.

3. Comparative Verification  
   Check for exact correspondence between the math and Lean statements; you may refer to aspects like:
   - Semantic alignment, logic structure, and quantifier correctness.
   - Preservation of constraints and boundary assumptions.
   - Accurate typing and use of variables.
   - Syntactic validity and proper Lean usage (free from errors).
   - Use of symbols and constructs without semantic drift.
   - No missing elements, no unjustified additions, and no automatic corrections or completions.

4. Final Judgement  
   Based solely on the above analysis, judge whether the Lean statement is a correct and exact formalization of the mathematical problem.

5. Accuracy Confirmation  
   If correct: clearly confirm why all elements match.  
   If incorrect: list all mismatches and explain how each one affects correctness.

Note: While the analysis may be broad and open to interpreting all relevant features, the final judgment must be based only on what is explicitly and formally expressed in the Lean statement.  
**Do not consider or assess any part of the proof. Your judgment should be entirely about the accuracy of the statement formalization.**

Output Format:
Please provide your analysis and conclusion. At the end of your response, clearly state your final judgment using one of these exact phrases:
- "Final Judgment: Correct" (if the formalization is correct)
- "Final Judgment: Incorrect" (if the formalization is incorrect)

You may format your response in any way you prefer (plain text, markdown, etc.), but make sure to include the "Final Judgment: Correct" or "Final Judgment: Incorrect" statement at the end.

Input Data:
— Start of Mathematical_Text —
{mathematical_statement}
— End of Mathematical_Text —

— Start of Lean4Code —
{autoformalization_placeholder}
— End of Lean4Code —
\end{Verbatim}
\end{resultbox}
\section{Error Analysis and Case Study}\label{appendix:error_analysis}

\subsection{Semantic Mismatch in Statement Formalization}

In this section, we present some failed cases in statement formalization due to semantic mismatch.

\subsubsection{DeepTheorem 665126}

\begin{resultbox}{Informal Statement}
Let $f: \text{Domain} \to \text{Range}$, $U$ be a special function of three variables, and $n \text{ and } k$ be integers such that $k \leq n$. If the function $\lambda$ is defined as: $$ \lambda = -\sum_{k=1}^{n} \sum_{c=1}^{\min(k,n-1)} {k \choose c} \frac{(-1)^k}{k!} f^{k-c}U(-c,k-2c+1,-f)\phi(n,k), $$ prove that the order of summation can be switched, and $\lambda$ can equivalently be expressed as: $$ \lambda = -\sum_{k=1}^{n} \frac{(-1)^k}{k!}\phi(n,k) \sum_{c=1}^{\min(k,n-1)} {k \choose c} f^{k-c}U(-c,k-2c+1,-f). $$
\end{resultbox}

\begin{codebox}{Formal Statement}
\begin{leancode}
theorem sum_rearrangement_proof (n : ℕ) (f : ℝ) (U : ℤ → ℤ → ℝ → ℝ) (phi : ℕ → ℕ → ℝ) :
  -∑ k in Finset.Icc 1 n, ∑ c in Finset.Icc 1 (min k (n - 1)), 
    (Nat.choose k c : ℝ) * ((-1 : ℝ) ^ k / (Nat.factorial k : ℝ)) * 
    f ^ (k - c) * U (-c) (k - 2 * c + 1) (-f) * phi n k = 
  -∑ k in Finset.Icc 1 n, 
    ((-1 : ℝ) ^ k / (Nat.factorial k : ℝ)) * phi n k * 
    (∑ c in Finset.Icc 1 (min k (n - 1)), 
      (Nat.choose k c : ℝ) * f ^ (k - c) * U (-c) (k - 2 * c + 1) (-f)) := by sorry
\end{leancode}
\end{codebox}

In the example above, the model makes a typical error involving natural number subtraction. 
Although the expression $k - 2c + 1$ is expected to be an integer based on the type of $U$, 
the subtraction itself is performed within the domain of natural numbers because $k, c \in \mathbb{N}$. 
Since $k - 2c + 1$ is not guaranteed to be positive, this leads to a result that diverges from 
the intended mathematical logic. In Lean, subtraction on the type \texttt{Nat} is 
\textit{truncated at zero} (i.e., $n - m = 0$ if $m > n$), a behavior known as 
\textit{saturating subtraction}.

\subsubsection{DeepTheorem 618039}
\begin{resultbox}{Informal Statement}
Let $y: \mathbb{R} \setminus \{0\} 	\to \mathbb{R}$ be a differentiable function satisfying the nonlinear first-order differential equation $y + x \frac{dy}{dx} = x^4 \left( \frac{dy}{dx} \right)^2$. Prove that the general solution is given by $$y = \frac{a}{x} + a^2,$$ where $a \in \mathbb{R}$ is a constant.
\end{resultbox}

\begin{codebox}{Formal Statement}
\begin{leancode}
theorem verify_differential_equation_solution (a : ℝ) : 
  let y : ℝ → ℝ := fun x => a / x + a^2
  (∀ x : ℝ, x ≠ 0 → DifferentiableAt ℝ y x) ∧
  (∀ x : ℝ, x ≠ 0 → y x + x * (deriv y x) = x^4 * (deriv y x)^2) := by sorry
\end{leancode}
\end{codebox}

In the example above, the model captures only a portion of the original proposition's semantics. 
Specifically, it verifies that $y = \frac{a}{x} + a^2$ is a solution to the differential equation, 
but fails to establish that all solutions must take this specific form. 
In other words, the model proves only the \textit{sufficiency} of the condition without addressing its \textit{necessity}.

\subsection{Failed Proof Sketch Analysis}

\begin{figure}[h]
    \centering
    \includegraphics[width=0.6\linewidth]{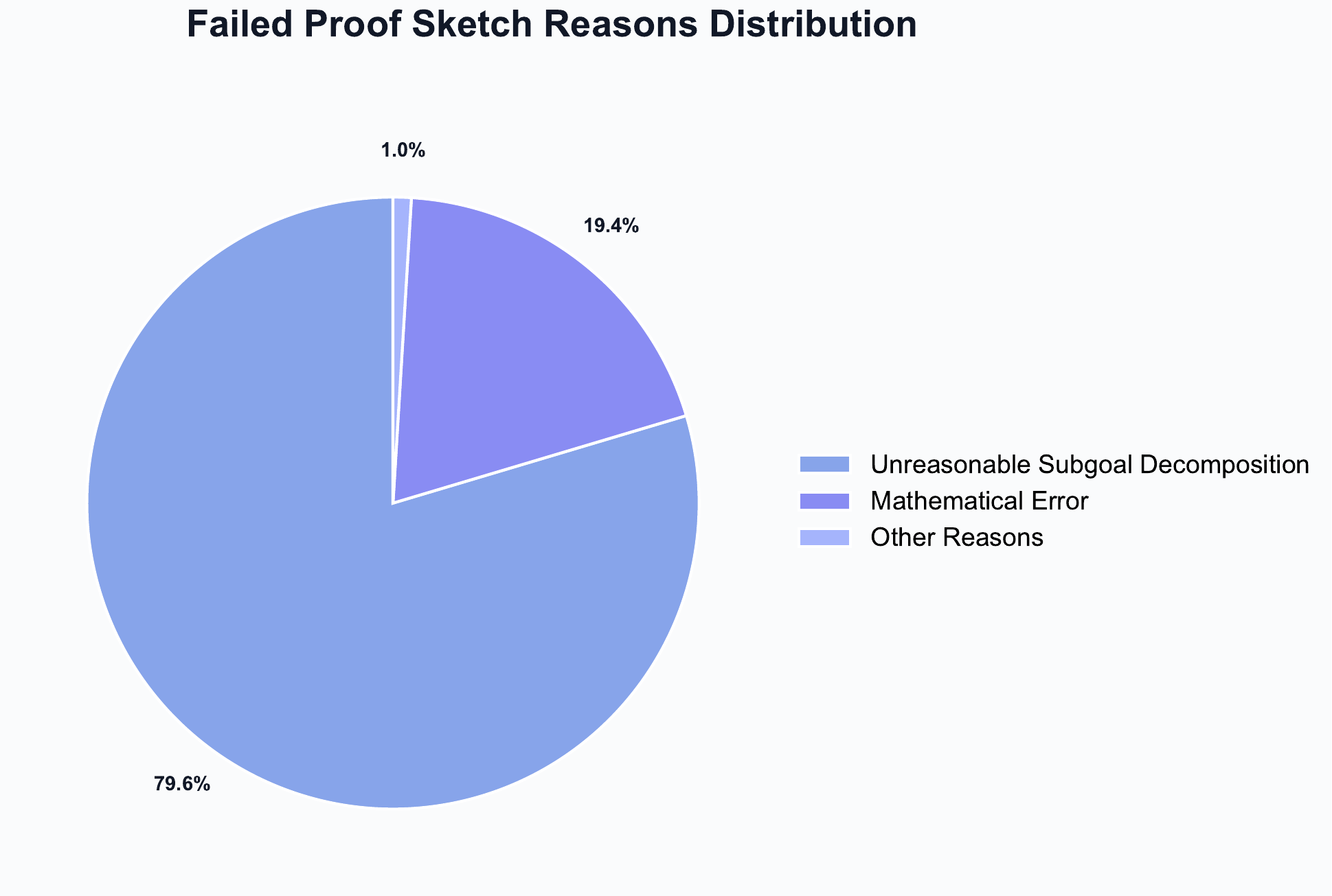}
    \caption{The failed reasons of the 520 sketches from large-scale benchmark run.}
    \label{fig:sketch_failed}
\end{figure}

To identify general failure patterns, we utilize Gemini-3-Flash to analyze the failed proof sketches and summarize the underlying causes. The results are presented in Figure~\ref{fig:sketch_failed}, where the errors are classified into four primary categories:
\begin{itemize}
\item \textbf{Unreasonable Subgoal Decomposition}: The model partitions the proof into subgoals that are logically disjointed or involve excessively large logical leaps that hinder formalization.
\item \textbf{Mathematical Error}: The sketch contains fundamental mathematical fallacies or involves subgoals that are logically false.
\item \textbf{Other}: Miscellaneous errors that do not fall into the above categories.
\end{itemize}

It should be noted that these results provide a general diagnostic trend rather than an absolute metric, given the potential for LLM hallucinations. As illustrated in Figure~\ref{fig:sketch_failed}, \textit{Unreasonable Subgoal Decomposition} (79.6\%) emerges as the overwhelming failure mode, representing the vast majority of unsuccessful sketches. The model primarily struggles with orchestrating a coherent and well-founded proof strategy. While such flawed sketches would be detrimental if naively included in training sets, they could serve as valuable negative samples for contrastive learning. Additionally, \textit{Mathematical Errors} (19.4\%) point to persistent imprecision during the informal-to-formal translation process. To gain a more granular understanding of these failure patterns, we conduct a manual analysis of representative cases in the following section.

\subsubsection{DeepMath 53832}
\begin{codebox}{Formal Statement}
\begin{leancode}
theorem count_distinct_terms_expansion : 
  ∃ (S : Finset ℤ), 
    (∀ k ∈ S, ∃ (a b c d : ℕ), 
      a + b + c + d = 15 ∧ 
      k = a - b + 2*c - 2*d) ∧
    (∀ k : ℤ, (∃ (a b c d : ℕ), 
      a + b + c + d = 15 ∧ 
      k = a - b + 2*c - 2*d) → k ∈ S) ∧
    S.card = 61 := by sorry
\end{leancode}
\end{codebox}

\begin{codebox}{Proof Sketch}
\begin{leancode}
theorem count_distinct_terms_expansion : 
  ∃ (S : Finset ℤ), 
    (∀ k ∈ S, ∃ (a b c d : ℕ), 
      a + b + c + d = 15 ∧ 
      k = a - b + 2*c - 2*d) ∧ 
    (∀ k : ℤ, (∃ (a b c d : ℕ), 
      a + b + c + d = 15 ∧ 
      k = a - b + 2*c - 2*d) → k ∈ S) ∧ 
    S.card = 61 := by
  -- Define the candidate set S as the interval of integers from -30 to 30
  let S : Finset ℤ := Finset.Icc (-30) 30
  use S
  -- Subgoal 1: Show that every k in the set [-30, 30] can be expressed in the required form
  have h_achievable : ∀ k ∈ S, ∃ (a b c d : ℕ), a + b + c + d = 15 ∧ k = a - b + 2 * c - 2 * d := by sorry
  -- Subgoal 2: Show that any k satisfying the equation must lie in the range [-30, 30]
  have h_bounded : ∀ k : ℤ, (∃ (a b c d : ℕ), a + b + c + d = 15 ∧ k = a - b + 2 * c - 2 * d) → k ∈ S := by sorry
  -- Subgoal 3: Prove that the maximum value of the expression is 30
  have h_max_val : (0 : ℤ) - 0 + 2 * 15 - 2 * 0 = 30 := by sorry
  -- Subgoal 4: Prove that the minimum value of the expression is -30
  have h_min_val : (0 : ℤ) - 0 + 2 * 0 - 2 * 15 = -30 := by sorry
  -- Subgoal 5: Calculate the cardinality of the integer interval Finset.Icc (-30) 30
  have h_card : S.card = 61 := by sorry
  -- Combine the subgoals to complete the proof
  exact ⟨h_achievable, h_bounded, h_card⟩
\end{leancode}
\end{codebox}

The aforementioned case serves as a representative example of unreasonable subgoal decomposition. While the partitioning of the proof into \texttt{h\_achievable}, \texttt{h\_bounded}, and \texttt{h\_card} is logically valid, given that the formal proposition is structured as a conjunction, it remains practically trivial. Such a decomposition does not facilitate meaningful mathematical progress. Specifically, although \texttt{h\_achievable} and \texttt{h\_bounded} are inherently complex, the generated sketch fails to provide any auxiliary sub-steps to assist in their derivation. This instance highlights the excessively coarse granularity of the current model's problem-solving approach. To address this limitation, we posit that an iterative subgoal decomposition strategy would be more effective in handling such complex formalization tasks.

\subsubsection{DeepMath 9543}
\begin{codebox}{Formal Statement}
\begin{leancode}
theorem count_even_sum_floor_div : 
  (Finset.filter (fun m => Even (∑ k in Finset.Icc 1 m, m / k)) (Finset.Icc 1 1000)).card = 495 := by sorry
\end{leancode}
\end{codebox}
\begin{codebox}{Proof Sketch}
\begin{leancode}
theorem count_even_sum_floor_div : 
  (Finset.filter (fun m => Even (∑ k in Finset.Icc 1 m, m / k)) (Finset.Icc 1 1000)).card = 495 := by
  have h1 : ∀ m ∈ Finset.Icc 1 1000, ∑ k in Finset.Icc 1 m, m / k = ∑ n in Finset.Icc 1 m, (Nat.divisors n).card := by sorry
  have h2 : ∀ n, Odd (Nat.divisors n).card ↔ IsSquare n := by sorry
  have h3 : ∀ m, (∑ n in Finset.Icc 1 m, (Nat.divisors n).card) \% 2 = (Finset.filter IsSquare (Finset.Icc 1 m)).card \% 2 := by sorry
  have h4 : ∀ m, (Finset.filter IsSquare (Finset.Icc 1 m)).card = Nat.sqrt m := by sorry
  have h5 : ∀ m ∈ Finset.Icc 1 1000, Even (∑ k in Finset.Icc 1 m, m / k) ↔ Even (Nat.sqrt m) := by sorry
  have h6 : (Finset.filter (fun m => Even (Nat.sqrt m)) (Finset.Icc 1 1000)).card = (Finset.filter (fun m => Even (Nat.sqrt m)) (Finset.Icc 1 1000)).card := by sorry
  have h7 : (Finset.filter (fun m => Even (Nat.sqrt m)) (Finset.Icc 1 1000)) = (Finset.Icc 1 15).biUnion (fun j => Finset.Icc (4 * j * j) (min 1000 (4 * j * j + 4 * j))) := by sorry
  have h8 : ∀ j ∈ Finset.Icc 1 15, (Finset.Icc (4 * j * j) (min 1000 (4 * j * j + 4 * j))).card = if j = 15 then 40 else 4 * j + 1 := by sorry
  have h9 : (Finset.filter (fun m => Even (Nat.sqrt m)) (Finset.Icc 1 1000)).card = (∑ j in Finset.Icc 1 14, (4 * j + 1)) + 40 := by sorry
  have h10 : (∑ j in Finset.Icc 1 14, (4 * j + 1)) + 40 = 4 * (∑ j in Finset.Icc 1 14, j) + 14 + 40 := by sorry
  have h11 : ∑ j in Finset.Icc 1 14, j = 105 := by sorry
  have h12 : 4 * 105 + 14 + 40 = 495 := by sorry
  rw [Finset.filter_congr h5]
  rw [h9]
  rw [h10]
  rw [h11]
  exact h12
\end{leancode}
\end{codebox}

In the aforementioned proof, a manifest mathematical fallacy occurs in \texttt{h8}: when $j=15$, the result should be 61 rather than 40. This initial error propagates through several subsequent subgoals, yet the final step, \texttt{h12}, ostensibly masks these underlying inconsistencies. Our inspection further reveals a significant discrepancy between the formal proof sketch and the original natural language proof. This mismatch stems partly from the inherent structural divergence between formal and informal mathematics, and partly from model hallucinations. Given that the failure of a single subgoal invalidates the entire formal verification, such hallucinations pose a severe bottleneck to the overall performance of the workflow.

Furthermore, we observed that a subset of failed sketches originates from flawed formal statements. These errors are attributed either to inherent inaccuracies within the source dataset or to semantic loss during the statement formalization process. These findings necessitate a more cautious evaluation of the proof sketches extracted from failed trajectories via the decoupled extraction strategy. Nevertheless, identifying these specific failure modes provides critical insights for the further optimization of our workflow.

\section{Data Example}\label{appendix:data_example}

\subsection{Statement Formalization}

\subsubsection{Example 1}
\begin{resultbox}{Informal Statement}
Find the Laplace Transform of the function $\delta(t-2\pi)\cos(t)$, where $\delta(t) $ is the Dirac Delta Function.

Answer: $e^{-2\pi s}$
\end{resultbox}

\begin{databox}{Premises}
\texttt{[SchwartzMap.delta, MeasureTheory.Measure.dirac, Real.cos, SchwartzMap.integralCLM\_dirac\_eq\_delta]}
\end{databox}

\begin{codebox}{Formal Statement}
\begin{leancode}
theorem laplace_transform_delta_cos :
  ∀ s : ℝ, (MeasureTheory.integral (MeasureTheory.Measure.dirac (2 * Real.pi)) 
    (fun t => Real.cos t * Real.exp (-s * t))) = 
        Real.exp (-2 * Real.pi * s) := by sorry
\end{leancode}
\end{codebox}

\subsubsection{Example 2}

\begin{resultbox}{Informal Statement}
Let $A$ be an $n \times n$ matrix over a field $\mathbf{F}$ and $b$ a vector in $\mathbf{F}^n$. Prove that if $A$ is invertible, then the linear system $A x = b$ has a unique solution $x$ in $\mathbf{F}^n$, given by $x = A^{-1} b$.
\end{resultbox}

\begin{databox}{Premises}
\texttt{[Matrix.inv, Invertible, Matrix.mulVec.addMonoidHomLeft, Matrix.GeneralLinearGroup, Matrix.termGL, Matrix.invertibleOfDetInvertible, Matrix.detInvertibleOfInvertible, Matrix.mul\_right\_inj\_of\_invertible]}
\end{databox}

\begin{codebox}{Formal Statement}
\begin{leancode}
theorem matrix_unique_solution {F : Type*} [Field F] {n : Type*} [Fintype n] [DecidableEq n]
  (A : Matrix n n F) (b : n → F) (hA : IsUnit A) :
  (∃! x : n → F, A.mulVec x = b) ∧ (∀ x, A.mulVec x = b ↔ x = A⁻¹.mulVec b) := by sorry
\end{leancode}
\end{codebox}

\subsection{Proof Generation}

\subsubsection{Example 1}
\begin{codebox}{Formal Statement}
\begin{leancode}
theorem h_trace_nilpotent (F : Type*) [Field F] (n : ℕ) (hn : 0 < n) (A : Matrix (Fin n) (Fin n) F) (h : IsNilpotent A) : IsNilpotent (Matrix.trace A) := by sorry
\end{leancode}    
\end{codebox}

\begin{databox}{Premises}
\texttt{[Matrix.isNilpotent\_trace\_of\_isNilpotent, IsNilpotent]}
\end{databox}

\begin{codebox}{Formal Proof}
\begin{leancode}
theorem h_trace_nilpotent (F : Type*) [Field F] (n : ℕ) (hn : 0 < n) (A : Matrix (Fin n) (Fin n) F) (h : IsNilpotent A) : IsNilpotent (Matrix.trace A) := by
  apply Matrix.isNilpotent_trace_of_isNilpotent h
\end{leancode}
\end{codebox}
\subsubsection{Example 2}

\begin{codebox}{Formal Statement}
\begin{leancode}
theorem trace_of_nilpotent_matrix (F : Type*) [Field F] (n : ℕ) (hn : 0 < n) (A : Matrix (Fin n) (Fin n) F) 
  (h : IsNilpotent A) : Matrix.trace A = 0 := by sorry
\end{leancode}
\end{codebox}

\begin{databox}{Premises}
\texttt{[Matrix.isNilpotent\_trace\_of\_isNilpotent, IsNilpotent]}
\end{databox}

\begin{codebox}{Formal Proof}
\begin{leancode}
theorem h_trace_nilpotent (F : Type*) [Field F] (n : ℕ) (hn : 0 < n) (A : Matrix (Fin n) (Fin n) F) (h : IsNilpotent A) : IsNilpotent (Matrix.trace A) := by
  apply Matrix.isNilpotent_trace_of_isNilpotent h
theorem h_field_nilpotent_is_zero (F : Type*) [Field F] {x : F} (hx : IsNilpotent x) : x = 0 := by
  have h_main : ∀ (n : ℕ), x ^ n = 0 → x = 0 := by
    intro n hn
    by_contra hx'
    -- If x ≠ 0, then x is invertible
    have h : x ≠ 0 := hx'
    -- Since x ≠ 0, we can find its inverse
    have h_inv : ∃ (y : F), x * y = 1 := by
      exact ⟨x⁻¹, by field_simp [h]⟩
    -- Obtain the inverse y such that x * y = 1
    obtain ⟨y, hy⟩ := h_inv
    -- Compute (x * y)^n = 1^n = 1
    have h1 : (x * y) ^ n = 1 := by
      calc
        (x * y) ^ n = 1 ^ n := by rw [hy]
        _ = 1 := by simp
    -- But x^n = 0, so (x * y)^n = 0 * y^n = 0
    have h2 : (x * y) ^ n = 0 := by
      calc
        (x * y) ^ n = x ^ n * y ^ n := by
          rw [mul_pow]
        _ = 0 * y ^ n := by rw [hn]
        _ = 0 := by simp
    -- Therefore, 1 = 0, which is a contradiction
    have h3 : (1 : F) = 0 := by
      calc
        (1 : F) = (x * y) ^ n := by rw [h1]
        _ = 0 := by rw [h2]
    -- Since 1 ≠ 0 in a field, we have a contradiction
    have h4 : (1 : F) ≠ 0 := by
      exact one_ne_zero
    -- Conclude that x must be 0
    exact h4 h3
  
  have h_final : x = 0 := by
    obtain ⟨n, hn⟩ := hx
    have h : x = 0 := h_main n hn
    exact h
  
  exact h_final
set_option maxHeartbeats 0

open BigOperators Real Nat Topology Rat

theorem trace_of_nilpotent_matrix (F : Type*) [Field F] (n : ℕ) (hn : 0 < n) (A : Matrix (Fin n) (Fin n) F) 
  (h : IsNilpotent A) : Matrix.trace A = 0 := by
  have h_trace_nilpotent : IsNilpotent (Matrix.trace A) := by
    exact h_trace_nilpotent F n hn A h
  have h_field_nilpotent_is_zero {x : F} (hx : IsNilpotent x) : x = 0 := by
    exact h_field_nilpotent_is_zero F hx
  exact h_field_nilpotent_is_zero h_trace_nilpotent
\end{leancode}
\end{codebox}

\subsection{Premise Selection}

\subsubsection{Example 1}

\begin{resultbox}{Informal Statement}
Determine if there exists a function $f(x)$ such that $f(a+b)-f(a)=kab+3b^2$ for all real numbers $a$ and $b$, given that $f(1)=3$ and $f(2)=10$. If such a function exists, find its derivative $f'(x)$. Provide your answer as an expression for $f'(x)$. 

Answer: 6x
\end{resultbox}

\begin{databox}{Queries}
\texttt{[properties of deriv in mathlib, definition of differentiable function on real numbers, derivative of a function f at point x]}
\end{databox}

\begin{databox}{Premises}
\texttt{[deriv, Differentiable]}
\end{databox}

\subsubsection{Example 2}

\begin{codebox}{Formal Statement}
\begin{leancode}
theorem not_cyclic_of_many_elements_of_order_p (G : Type*) [Group G] [Fintype G] (p : ℕ) (hp : Nat.Prime p) 
  (h : Set.ncard {x : G | orderOf x = p} > p - 1) : 
  ¬ IsCyclic G := by sorry
\end{leancode}
\end{codebox}

\begin{databox}{Queries}
\texttt{[phi\_prime, ncard of set of elements with specific order in cyclic group, card\_orderOf\_eq\_totient, number of elements of order d in a cyclic group, isCyclic\_iff\_exists\_orderOf\_eq\_card]}
\end{databox}

\begin{databox}{Premises}
\texttt{[IsCyclic.card\_orderOf\_eq\_totient, Nat.totient\_prime, Prime]}
\end{databox}

\subsection{Proof Correction}

\subsubsection{Example 1}

\begin{codebox}{Failed Code}
\begin{leancode}
theorem h_map_eval (P : Polynomial ℤ)
  (h_eval_zero : P.eval 1 = 0) :
  (P.map (algebraMap ℤ ℂ)).eval 1 = algebraMap ℤ ℂ (P.eval 1) := by
  -- We want to show that evaluating the mapped polynomial at 1
  -- is the same as mapping the evaluation of the original polynomial at 1.
  -- This is a general property of polynomial maps and evaluation.
  -- The theorem `Polynomial.eval_map` states: (p.map f).eval (f x) = f (p.eval x)
  -- Here, f = algebraMap ℤ ℂ and x = 1.
  -- Note that (algebraMap ℤ ℂ) 1 = 1.
  have h_one : (algebraMap ℤ ℂ) 1 = 1 := by simp
  rw [← h_one, Polynomial.eval_map]
\end{leancode}
\end{codebox}

\begin{databox}{Error Messages}
\begin{leancode}
Error 1:

Corresponding Code:
```lean4
theorem h_map_eval (P : Polynomial ℤ)
  (h_eval_zero : P.eval 1 = 0) :
  (P.map (algebraMap ℤ ℂ)).eval 1 = algebraMap ℤ ℂ (P.eval 1) := <error>by
  -- We want to show that evaluating the mapped polynomial at 1
  -- is the same as mapping the evaluation of the original polynomial at 1.
  -- This is a general property of polynomial maps and evaluation.
  -- The theorem `Polynomial.eval_map` states: (p.map f).eval (f x) = f (p.eval x)
  -- Here, f = algebraMap ℤ ℂ and x = 1.
  -- Note that (algebraMap ℤ ℂ) 1 = 1.
  have h_one : (algebraMap ℤ ℂ) 1 = 1 := by simp
  rw [← h_one, Polynomial.eval_map]</error>

```

Error Message: unsolved goals
P : Polynomial ℤ
h_eval_zero : Polynomial.eval 1 P = 0
h_one : (algebraMap ℤ ℂ) 1 = 1
⊢ Polynomial.eval₂ (algebraMap ℤ ℂ) ((algebraMap ℤ ℂ) 1) P = (algebraMap ℤ ℂ) (Polynomial.eval 1 P)
\end{leancode}
\end{databox}

\begin{codebox}{Valid Code}
\begin{leancode}
theorem h_map_eval (P : Polynomial ℤ)
  (h_eval_zero : P.eval 1 = 0) :
  (P.map (algebraMap ℤ ℂ)).eval 1 = algebraMap ℤ ℂ (P.eval 1) := by
  -- The core property here is that evaluation of a mapped polynomial at f(x)
  -- is the same as applying f to the evaluation of the original polynomial at x.
  -- This is stated in the lemma `eval_map`.
  -- Since algebraMap ℤ ℂ is a ring homomorphism, it preserves 1.
  have h_one : (algebraMap ℤ ℂ) 1 = 1 := by simp
  rw [← h_one]
  -- Now we can use the theorem: (P.map f).eval (f x) = f (P.eval x)
  -- In mathlib, this is exactly `Polynomial.eval_map` combined with some simplification.
  -- Or more directly, `hom_eval` or simply `eval_map`.
  -- Let's check the behavior of eval_map: (p.map f).eval x = p.eval₂ f x.
  -- There is another identity: eval₂ f (f x) p = f (eval x p).
  -- This identity is provided by `Polynomial.eval₂_at_apply`.
  rw [Polynomial.eval_map]
  exact Polynomial.eval₂_at_apply (algebraMap ℤ ℂ) 1
\end{leancode}
\end{codebox}

\subsubsection{Example 2}

\begin{codebox}{Failed Code}
\begin{leancode}
theorem h_exp_div_lim : Filter.Tendsto (fun u : ℝ => Real.exp u / u) Filter.atTop Filter.atTop := by
  -- We use the theorem tendsto_exp_div_rpow_atTop which states that 
  -- exp(x) / x^s tends to +∞ at +∞ for any real s.
  -- Here, our function is exp(u) / u, which is equivalent to exp(u) / u^1.
  -- The theorem is available as tendsto_exp_div_rpow_atTop.
  have h := tendsto_exp_div_rpow_atTop 1
  -- Since x^1 = x for x : ℝ, we can use simp to transform the expression.
  -- In Lean 4, x^1 is simplified to x for real numbers.
  simp only [pow_one] at h
  exact h
\end{leancode}
\end{codebox}

\begin{databox}{Error Messages}
\begin{leancode}
Error 1:

Corresponding Code:
```lean4
  have h := tendsto_exp_div_rpow_atTop 1
  -- Since x^1 = x for x : ℝ, we can use simp to transform the expression.
  -- In Lean 4, x^1 is simplified to x for real numbers.
  simp only [pow_one] at h
  <error>exact h</error>

```

Error Message: type mismatch
  h
has type
  Filter.Tendsto (fun x => Real.exp x / x ^ 1) Filter.atTop Filter.atTop : Prop
but is expected to have type
  Filter.Tendsto (fun u => Real.exp u / u) Filter.atTop Filter.atTop : Prop
\end{leancode}
\end{databox}

\begin{codebox}{Valid Code}
\begin{leancode}
theorem h_exp_div_lim : Filter.Tendsto (fun u : ℝ => Real.exp u / u) Filter.atTop Filter.atTop := by
  -- Use the existing theorem that states exp(x) / x^s tends to +∞ as x tends to +∞.
  -- Here s = 1.
  have h := tendsto_exp_div_rpow_atTop 1
  -- In Lean 4, for real numbers x, the expression x ^ (1 : ℝ) is the same as x.
  -- The theorem tendsto_exp_div_rpow_atTop uses the real power function (rpow).
  -- We use Real.rpow_one to simplify x^1 to x.
  simp only [Real.rpow_one] at h
  exact h
\end{leancode}
\end{codebox}

\subsection{Proof Sketching}

\subsubsection{Example 1}

\begin{codebox}{Formal Statement}
\begin{leancode}
theorem polynomial_symmetric_odd_coefficients_has_unit_root (n : ℕ) (P : Polynomial ℤ) 
  (h_degree : P.natDegree = n)
  (h_symm : ∀ i : ℕ, i ≤ n → P.coeff i = -(P.coeff (n - i)))
  (h_odd : ∀ i : ℕ, i ≤ n → Odd (P.coeff i)) :
  ∃ z : ℂ, Complex.abs z = 1 ∧ (P.map (algebraMap ℤ ℂ)).eval z = 0 := by sorry
\end{leancode}
\end{codebox}

\begin{resultbox}{Informal Proof}
1.  Let $P(x) = \sum_{i=0}^n a_i x^i$ be a polynomial with integer coefficients where $a_i$ denotes the coefficient of $x^i$ (given by $P.coeff(i)$).

2.  We are given that $n$ is the natural degree of $P$, so $a_n \neq 0$.

3.  We are given the symmetry condition: for all $i \in \{0, \dots, n\}$, $a_i = -a_{n-i}$.

4.  We are given the parity condition: for all $i \in \{0, \dots, n\}$, $a_i$ is an odd integer.

5.  Consider the symmetry condition specifically for the "middle" coefficient when $n$ is even. If $n = 2k$ for some integer $k$, then for $i = k$, we have $a_k = -a_{n-k} = -a_{2k-k} = -a_k$.

6.  This implies $2a_k = 0$, so $a_k = 0$.

7.  However, the parity condition states that $a_k$ must be an odd integer. By the theorem `Int.not\_odd\_zero`, $0$ is not odd.

8.  Thus, the assumption that $n$ is even leads to a contradiction. Therefore, $n$ must be an odd natural number.

9.  Now, let's evaluate the polynomial $P$ at $x = 1$. By the theorem `Polynomial.eval\_eq\_sum`, the evaluation of $P$ at $1$ is the sum of its coefficients: $P(1) = \sum_{i=0}^n a_i$.

10. Using the symmetry condition $a_i = -a_{n-i}$, we can rewrite the sum:
    $P(1) = \sum_{i=0}^n (-a_{n-i})$.
    
11. By performing a change of indices $j = n - i$, as $i$ goes from $0$ to $n$, $j$ also goes from $0$ to $n$: $P(1) = -\sum_{j=0}^n a_j$.

12. This shows that $P(1) = -P(1)$, which implies $2P(1) = 0$, and thus $P(1) = 0$ in the ring of integers $\mathbb{Z}$.

13. Now, we map the polynomial $P$ from $\mathbb{Z}[X]$ to $\mathbb{C}[X]$ using the canonical homomorphism `algebraMap $\mathbb{Z}$ $\mathbb{C}$`. Let $P_\mathbb{C} = P.map(algebraMap \mathbb{Z} \mathbb{C})$.

14. Evaluation is preserved under the ring homomorphism, so $P_\mathbb{C}(1) = (P(1))_\mathbb{C} = 0_\mathbb{C} = 0$.

15. We are looking for a complex number $z$ such that $|z| = 1$ and $P_\mathbb{C}(z) = 0$.

16. Let $z = 1$. The complex absolute value of $1$ is $|1| = 1$.

17. As shown in step 14, $P_\mathbb{C}(1) = 0$.

18. Thus, $z = 1$ is a root of the polynomial $P_\mathbb{C}$ with $|z| = 1$.

19. This concludes the proof that there exists a $z \in \mathbb{C}$ such that $|z| = 1$ and $P_\mathbb{C}(z) = 0$.
\end{resultbox}

\begin{databox}{Premises}
\texttt{[Polynomial.eval\_eq\_sum, Int.not\_odd\_zero, Polynomial.sum, Polynomial.sum\_monomial\_eq, Polynomial.eval\_sum, Polynomial.eval\textsubscript{2}\_eq\_sum]}
\end{databox}

\begin{codebox}{Proof Sketch}
\begin{leancode}
theorem polynomial_symmetric_odd_coefficients_has_unit_root (n : ℕ) (P : Polynomial ℤ) 
  (h_degree : P.natDegree = n)
  (h_symm : ∀ i : ℕ, i ≤ n → P.coeff i = -(P.coeff (n - i)))
  (h_odd : ∀ i : ℕ, i ≤ n → Odd (P.coeff i)) :
  ∃ z : ℂ, Complex.abs z = 1 ∧ (P.map (algebraMap ℤ ℂ)).eval z = 0 := by
  have h_eval_sum : P.eval 1 = ∑ i in Finset.range (n + 1), P.coeff i := by sorry
  have h_symm_sum : ∑ i in Finset.range (n + 1), P.coeff i = ∑ i in Finset.range (n + 1), -P.coeff (n - i) := by sorry
  have h_neg_sum : ∑ i in Finset.range (n + 1), -P.coeff (n - i) = - ∑ i in Finset.range (n + 1), P.coeff (n - i) := by sorry
  have h_reindex : ∑ i in Finset.range (n + 1), P.coeff (n - i) = ∑ i in Finset.range (n + 1), P.coeff i := by sorry
  have h_eval_eq_neg_eval : P.eval 1 = -P.eval 1 := by sorry
  have h_eval_zero : P.eval 1 = 0 := by sorry
  have h_map_eval : (P.map (algebraMap ℤ ℂ)).eval 1 = algebraMap ℤ ℂ (P.eval 1) := by sorry
  have h_map_eval_zero : (P.map (algebraMap ℤ ℂ)).eval 1 = 0 := by sorry
  have h_abs_one : Complex.abs 1 = 1 := by sorry
  refine ⟨1, h_abs_one, h_map_eval_zero⟩
\end{leancode} 
\end{codebox}

\subsubsection{Example 2}

\begin{codebox}{Formal Statement}
\begin{leancode}
theorem minimize_expression_with_sqrt_constraint : 
  (∃ x y : ℝ, 0 < x ∧ x < 2 ∧ y > 0 ∧ 
    (x - y)^2 + (Real.sqrt (4 - x^2) - 16 / y)^2 = 36 - 16 * Real.sqrt 2) ∧
  (∀ x y : ℝ, 0 < x ∧ x < 2 ∧ y > 0 → 
    (x - y)^2 + (Real.sqrt (4 - x^2) - 16 / y)^2 ≥ 36 - 16 * Real.sqrt 2) := by sorry
\end{leancode}
\end{codebox}

\begin{resultbox}{Informal Proof}
1. Let $P_1 = (x, \sqrt{4 - x^2})$ where $0 < x < 2$. Since $x^2 + (\sqrt{4 - x^2})^2 = 4$ and $x > 0, \sqrt{4 - x^2} > 0$, $P_1$ is a point on the arc of the circle $x^2 + y^2 = 4$ in the first quadrant.

2. Let $P_2 = (y, \frac{16}{y})$ where $y > 0$. Since $y \cdot \frac{16}{y} = 16$, $P_2$ is a point on the branch of the hyperbola $xy = 16$ in the first quadrant.

3. The expression $(x - y)^2 + (\sqrt{4 - x^2} - 16/y)^2$ is the square of the Euclidean distance between $P_1$ and $P_2$, denoted $d(P_1, P_2)^2$.

4. Both the circle $x^2 + y^2 = 4$ and the hyperbola $xy = 16$ are symmetric with respect to the line $L: y = x$.

5. Let $P_1^* = (\sqrt{2}, \sqrt{2})$. Since $(\sqrt{2})^2 + (\sqrt{2})^2 = 2 + 2 = 4$, $P_1^*$ is on the circle. This corresponds to setting $x = \sqrt{2}$ in the expression. Note $0 < \sqrt{2} < 2$.

6. Let $P_2^* = (4, 4)$. Since $4 \cdot 4 = 16$, $P_2^*$ is on the hyperbola. This corresponds to setting $y = 4$ in the expression. Note $4 > 0$.

7. Calculate the squared distance at these points: $d(P_1^*, P_2^*)^2 = (\sqrt{2} - 4)^2 + (\sqrt{2} - 4)^2 = 2(\sqrt{2} - 4)^2 = 2(2 - 8\sqrt{2} + 16) = 2(18 - 8\sqrt{2}) = 36 - 16\sqrt{2}$. This proves the first part of the theorem: the value $36 - 16\sqrt{2}$ is attained.

8. To show this is the minimum, we use the property that the minimum distance between two convex sets (or smooth curves) is realized along a common normal.

9. For the circle $g_1(x, y) = x^2 + y^2 - 4 = 0$, the gradient is $\nabla g_1 = (2x, 2y)$. At $P_1^* = (\sqrt{2}, \sqrt{2})$, $\nabla g_1 = (2\sqrt{2}, 2\sqrt{2})$, which is parallel to the vector $(1, 1)$. Thus the normal line to the circle at $P_1^*$ is the line $y = x$.

10. For the hyperbola $g_2(x, y) = xy - 16 = 0$, the gradient is $\nabla g_2 = (y, x)$. At $P_2^* = (4, 4)$, $\nabla g_2 = (4, 4)$, which is also parallel to the vector $(1, 1)$. Thus the normal line to the hyperbola at $P_2^*$ is also the line $y = x$.

11. The points $P_1^*$ and $P_2^*$ both lie on the line $y = x$, and their normals are aligned along this line.

12. Consider the tangent line to the circle at $P_1^*$, which is $x + y = 2\sqrt{2}$. Since the circle is centered at the origin, the circle lies in the region $x + y \le 2\sqrt{2}$ (specifically, the part of the circle in the first quadrant).

13. Consider the tangent line to the hyperbola at $P_2^*$, which is $x + y = 8$. Because $xy = 16$ is a convex function in the first quadrant, the hyperbola lies in the region $x + y \ge 8$.

14. For any point $P_1$ on the circle and $P_2$ on the hyperbola in the first quadrant: $d(P_1, P_2) \ge \text{dist}(P_1, \text{Line } x+y=8) + \text{dist}(\text{Line } x+y=8, \text{Line } x+y=2\sqrt{2}) + \text{dist}(\text{Line } x+y=2\sqrt{2}, P_2)$ is not quite right, but we can use the projection onto the line $y = x$.

15. Let $u = \frac{1}{\sqrt{2}}(1, 1)$ be the unit vector along $y=x$. For any $P_1 = (x_1, y_1)$ on the circle arc, the projection onto $u$ is $P_1 \cdot u = \frac{x_1 + y_1}{\sqrt{2}}$. Since $x_1^2 + y_1^2 = 4$ and $x_1, y_1 > 0$, by Cauchy-Schwarz, $x_1 + y_1 \le \sqrt{1^2+1^2}\sqrt{x_1^2+y_1^2} = \sqrt{2} \cdot 2 = 2\sqrt{2}$. Thus $P_1 \cdot u \le 2$.

16. For any $P_2 = (x_2, y_2)$ on the hyperbola branch, the projection onto $u$ is $P_2 \cdot u = \frac{x_2 + y_2}{\sqrt{2}}$. Since $x_2 y_2 = 16$ and $x_2, y_2 > 0$, by AM-GM, $x_2 + y_2 \ge 2\sqrt{x_2 y_2} = 2\sqrt{16} = 8$. Thus $P_2 \cdot u \ge \frac{8}{\sqrt{2}} = 4\sqrt{2}$.

17. The distance $d(P_1, P_2)$ is at least the distance between their projections on the line $y=x$: $d(P_1, P_2) = \|P_1 - P_2\| \ge |P_2 \cdot u - P_1 \cdot u| \ge 4\sqrt{2} - 2$.

18. Squaring both sides: $d(P_1, P_2)^2 \ge (4\sqrt{2} - 2)^2 = 32 - 16\sqrt{2} + 4 = 36 - 16\sqrt{2}$.

19. This confirms that for all $0 < x < 2$ and $y > 0$, $(x - y)^2 + (\sqrt{4 - x^2} - 16/y)^2 \ge 36 - 16\sqrt{2}$.
\end{resultbox}

\begin{databox}{Premises}
\texttt{[EuclideanSpace.dist\_eq, Euclidean.dist, real\_inner\_le\_norm', isLocalMin\_of\_deriv\_Ioo, norm\_inner\_le\_norm, abs\_real\_inner\_le\_norm, InnerProductSpace.Core.norm\_inner\_le\_norm, isLocalMin\_of\_deriv', inner\_mul\_inner\_self\_le]}
\end{databox}

\begin{codebox}{Proof Sketch}
\begin{leancode}
theorem minimize_expression_with_sqrt_constraint : 
  (∃ x y : ℝ, 0 < x ∧ x < 2 ∧ y > 0 ∧ 
    (x - y)^2 + (Real.sqrt (4 - x^2) - 16 / y)^2 = 36 - 16 * Real.sqrt 2) ∧
  (∀ x y : ℝ, 0 < x ∧ x < 2 ∧ y > 0 → 
    (x - y)^2 + (Real.sqrt (4 - x^2) - 16 / y)^2 ≥ 36 - 16 * Real.sqrt 2) := by
  have h_exists : ∃ x y : ℝ, 0 < x ∧ x < 2 ∧ y > 0 ∧ (x - y)^2 + (Real.sqrt (4 - x^2) - 16 / y)^2 = 36 - 16 * Real.sqrt 2 := by
    have h_val_x : Real.sqrt 2 > 0 ∧ Real.sqrt 2 < 2 := by sorry
    have h_val_y : (4 : ℝ) > 0 := by sorry
    have h_calc : (Real.sqrt 2 - 4)^2 + (Real.sqrt (4 - (Real.sqrt 2)^2) - 16 / 4)^2 = 36 - 16 * Real.sqrt 2 := by sorry
    exact ⟨Real.sqrt 2, 4, h_val_x.1, h_val_x.2, h_val_y, h_calc⟩
  have h_min : ∀ x y : ℝ, 0 < x ∧ x < 2 ∧ y > 0 → (x - y)^2 + (Real.sqrt (4 - x^2) - 16 / y)^2 ≥ 36 - 16 * Real.sqrt 2 := by
    intro x y h
    have hx_low : 0 < x := by sorry
    have hx_high : x < 2 := by sorry
    have hy_pos : y > 0 := by sorry
    let P1 : ℝ × ℝ := (x, Real.sqrt (4 - x^2))
    let P2 : ℝ × ℝ := (y, 16 / y)
    have h_dist_sq : (x - y)^2 + (Real.sqrt (4 - x^2) - 16 / y)^2 = (P2.1 - P1.1)^2 + (P2.2 - P1.2)^2 := by sorry
    have h_unit_vec : (P2.1 - P1.1)^2 + (P2.2 - P1.2)^2 ≥ (1/2 : ℝ) * ((P2.1 + P2.2) - (P1.1 + P1.2))^2 := by sorry
    have h_circle_bound : x + Real.sqrt (4 - x^2) ≤ 2 * Real.sqrt 2 := by sorry
    have h_hyperbola_bound : y + 16 / y ≥ 8 := by sorry
    have h_diff_bound : (y + 16 / y) - (x + Real.sqrt (4 - x^2)) ≥ 8 - 2 * Real.sqrt 2 := by sorry
    have h_sq_bound : (1/2 : ℝ) * ((y + 16 / y) - (x + Real.sqrt (4 - x^2)))^2 ≥ (1/2 : ℝ) * (8 - 2 * Real.sqrt 2)^2 := by sorry
    have h_final_calc : (1/2 : ℝ) * (8 - 2 * Real.sqrt 2)^2 = 36 - 16 * Real.sqrt 2 := by sorry
    calc (x - y)^2 + (Real.sqrt (4 - x^2) - 16 / y)^2 
      _ = (P2.1 - P1.1)^2 + (P2.2 - P1.2)^2 := h_dist_sq
      _ ≥ (1/2) * ((P2.1 + P2.2) - (P1.1 + P1.2))^2 := h_unit_vec
      _ ≥ 36 - 16 * Real.sqrt 2 := by sorry
  exact ⟨h_exists, h_min⟩
\end{leancode}
\end{codebox}

\include{refs}

\end{document}